\documentclass[letterpaper,10pt,journal]{IEEEtran}
\usepackage{amsmath,amsfonts}
\usepackage{algorithmic}
\usepackage{algorithm}
\usepackage{array}

\usepackage{textcomp}
\usepackage{stfloats}
\usepackage{url}
\usepackage{verbatim}
\usepackage{graphicx}
\usepackage{cite}
\usepackage[T1]{fontenc}
\usepackage{newtxtext,newtxmath}
\usepackage{float}
\usepackage{subcaption}
\usepackage{xcolor}
\usepackage{array}
\usepackage{colortbl}
\usepackage{booktabs}
\usepackage{multirow}
\usepackage{hyperref}
\begin{document}

\title{Learning to Recover: Dynamic Reward Shaping with Wheel-Leg Coordination for Fallen Robots}

\author{Boyuan Deng$^{1,2,3*}$, Luca Rossini$^{1}$, Jin Wang$^{1}$, Weijie Wang$^{1}$, Dimitrios Kanoulas$^{3}$, and Nikolaos Tsagarakis$^{1}$% 
% \thanks{*This work was not supported by any organization}% <-this % stops a space
\thanks{$^{1}$Humanoids and Human-Centered Mechatronics (HHCM), Istituto Italiano di Tecnologia, Via Morego 30, Genoa, 16163, Italy}%
\thanks{$^{2}$Ph.D. Program of National Interest in Robotics and Intelligent Machines (DRIM), University of Genova, 16126 Genoa, Italy}
\thanks{$^{3}$ Department of Computer Science, University College London, Gower Street, WC1E 6BT, London, UK.}
\thanks{$^{*}$Corresponding author: boyuan.deng@iit.it}
\thanks{This work was supported by the UKRI FLF [MR/V025333/1] (RoboHike).  For the purpose of Open Access, the author has applied a CC BY public copyright license to any Author Accepted Manuscript version arising from this submission.}}



\maketitle

\begin{abstract}
Adaptive fall recovery is a critical capability for the practical deployment of wheeled-legged robots, which uniquely combine the agility of legs with the speed of wheels for rapid and efficient recovery. However, conventional approaches that depend on preplanned motions, simplified dynamics, or sparse rewards often fail to yield robust recovery behaviors. This paper introduces a learning-based framework that unites episode-based dynamic reward shaping with curriculum learning to balance the exploration of diverse recovery maneuvers and the refinement of precise postures. An asymmetric actor critic architecture accelerates training by leveraging privileged simulation information, while noise-injected observations improve robustness under uncertainty.
We further show that synergistic wheel–leg coordination reduces joint torque consumption by $15.8\%$ and $ 26.2\%$ and improves stabilization through energy transfer mechanisms. Extensive evaluations on two distinct quadruped platforms achieve recovery success rates up to $99.1\%$ and $97.8\%$ without platform-specific tuning. Supplementary material is available at \url{https://anonymous.4open.science/w/L2R-WheelLegCoordination-1A73/}
%Adaptive recovery from falls are essential skills for the practical deployment of wheeled-legged robots, which uniquely combine the agility of legs with the speed of wheels for rapid recovery. However, traditional methods that rely on preplanned recovery motions, simplified to dynamics, or sparse rewards often fail to produce robust recovery policies. This paper presents a learning-based framework that integrates Episode-based Dynamic Reward Shaping and curriculum learning, which dynamically balances exploration of diverse recovery maneuvers with precise posture refinement. An asymmetric actor-critic architecture accelerates training by leveraging privileged information in simulation, while noise-injected observations enhance robustness against uncertainties. We further demonstrate that synergistic wheel-leg coordination reduces joint torque consumption by $15.8\%$ and $ 26.2\%$ and improves stabilization through energy transfer mechanisms. Extensive evaluations on two distinct quadruped platforms achieve recovery success rates up to $99.1\%$ and $97.8\%$ without platform-specific tuning. Supplementary material is available at \url{https://anonymous.4open.science/w/L2R-WheelLegCoordination-1A73/}
\end{abstract}

\begin{IEEEkeywords}
Post-Fall Recovery, Dynamic Reward Shaping, Wheeled-Legged Robots, Wheel–Leg Coordination.
\end{IEEEkeywords}

\section{Introduction}
\IEEEPARstart{W}{heeled}-legged robots, with their integrated wheel–leg structure, have demonstrated unique mobility advantages in complex environments, establishing them as key platforms for long-duration tasks such as inspection and exploration\cite{bellicoso2018advances,bjelonic2022survey,bouman2020autonomous,tranzatto2022cerberus}. However, unexpected falls resulting from dynamic disturbances (e.g. collisions and slippage) often interrupt missions, and existing systems typically rely on manual intervention for recovery or preplanned recovery motions, which, while they can be effective on flat terrains, are not robust against diverse terrain geometries, severely limiting operational efficiency and autonomy. Thus, breakthroughs in robust and autonomous post-fall recovery are critical to achieving fully autonomous missions \cite{christensen2021roadmap}.

In essence, recovery after a fall is a highly complex motion planning and control problem, necessitating precise posture adjustments through multiple discontinuous contact events\cite{ma2023learning}, while the ground specific geometry can impose additional contact uncertainty challenges, which can compromise the successful execution of the fall recovery actions. For the controller, deriving an optimal or near-optimal sequence of actions is highly challenging. Due to the highly nonlinear dynamics of wheeled-legged robots, optimization-based methods often depend on simplified system models and precise state estimation. Consequently, most of the existing work has focused on pure legged recovery\cite{castano2019design} mainly demonstrated on flat terrains, leaving the full potential of wheel–leg collaborative recovery largely unexplored. In contrast, recent advances in reinforcement learning have demonstrated remarkable progress in addressing this issue, offering a new technical pathway toward more flexible and robust recovery strategies.
\begin{figure}[t]
    \centering
    \begin{subfigure}[b]{0.15\textwidth}
        \centering
        \includegraphics[width=\textwidth]{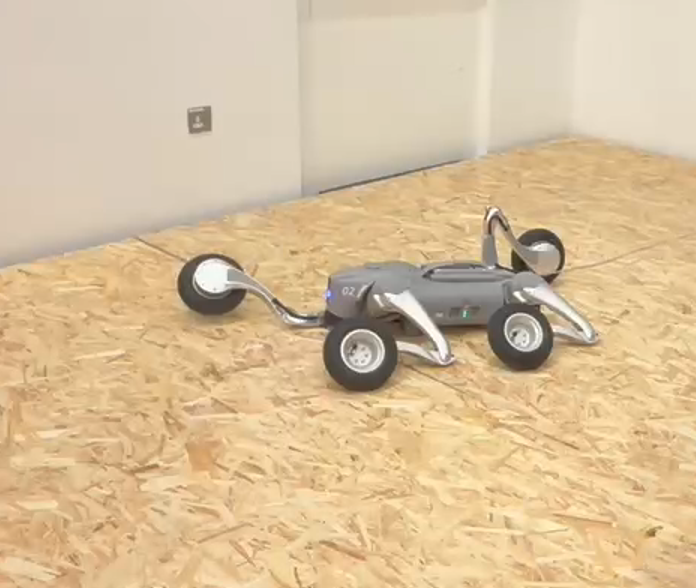}
        \caption{}
    \end{subfigure}
    % \hspace{1pt}
    \begin{subfigure}[b]{0.15\textwidth}
        \centering
        \includegraphics[width=\textwidth]{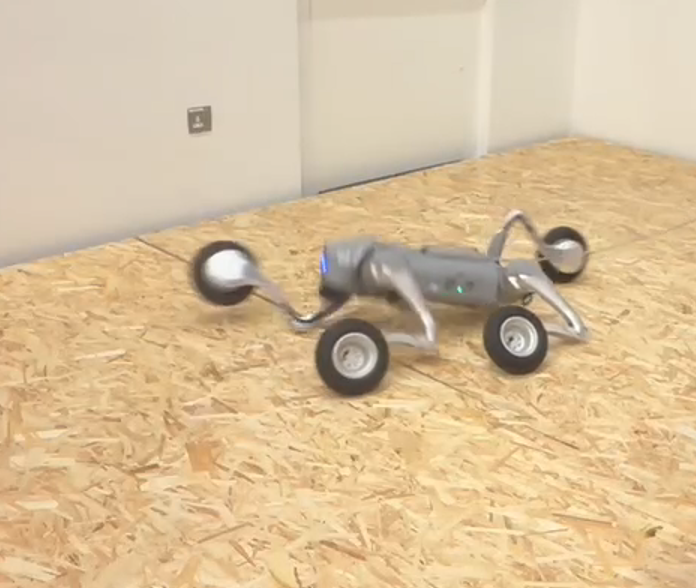}
        \caption{}
    \end{subfigure}
    % \hspace{1pt}
    \begin{subfigure}[b]{0.15\textwidth}
        \centering
        \includegraphics[width=\textwidth]{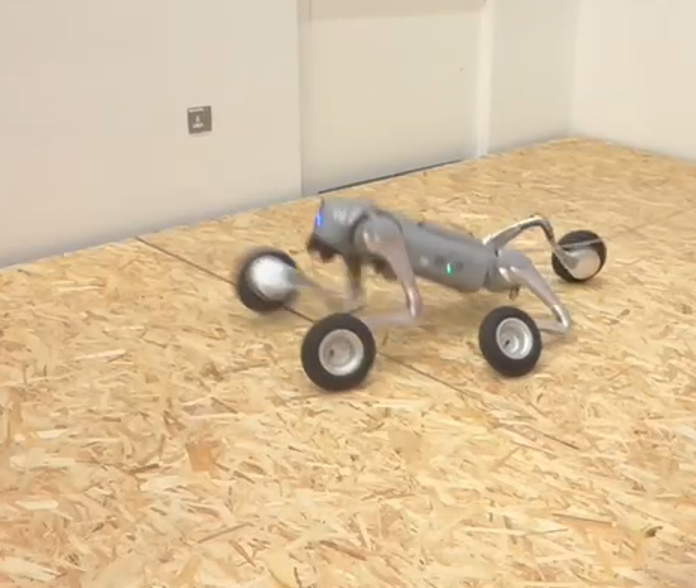}
        \caption{}
    \end{subfigure}
    % \vspace{0.1pt}
    \\
    \begin{subfigure}[b]{0.15\textwidth}
        \centering
        \includegraphics[width=\textwidth]{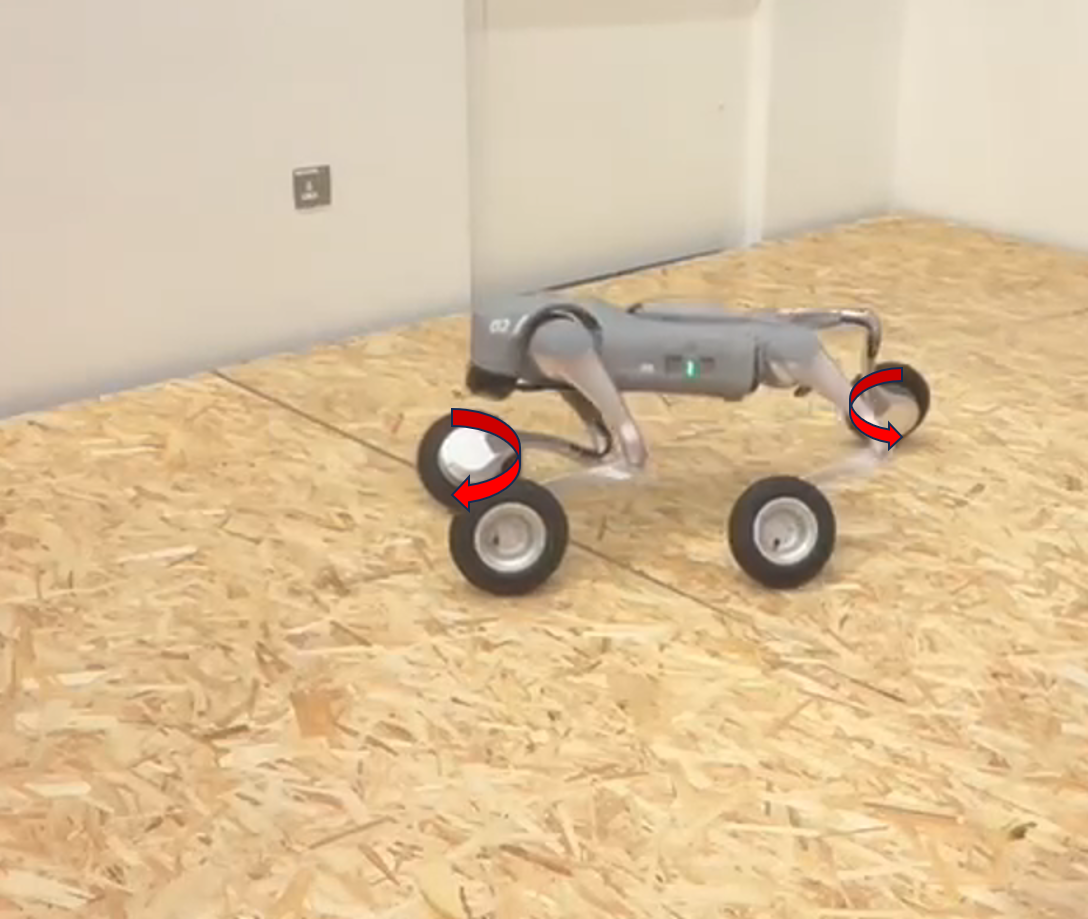}
        \caption{}
    \end{subfigure}
    % \hspace{1pt}
    \begin{subfigure}[b]{0.15\textwidth}
        \centering
        \includegraphics[width=\textwidth]{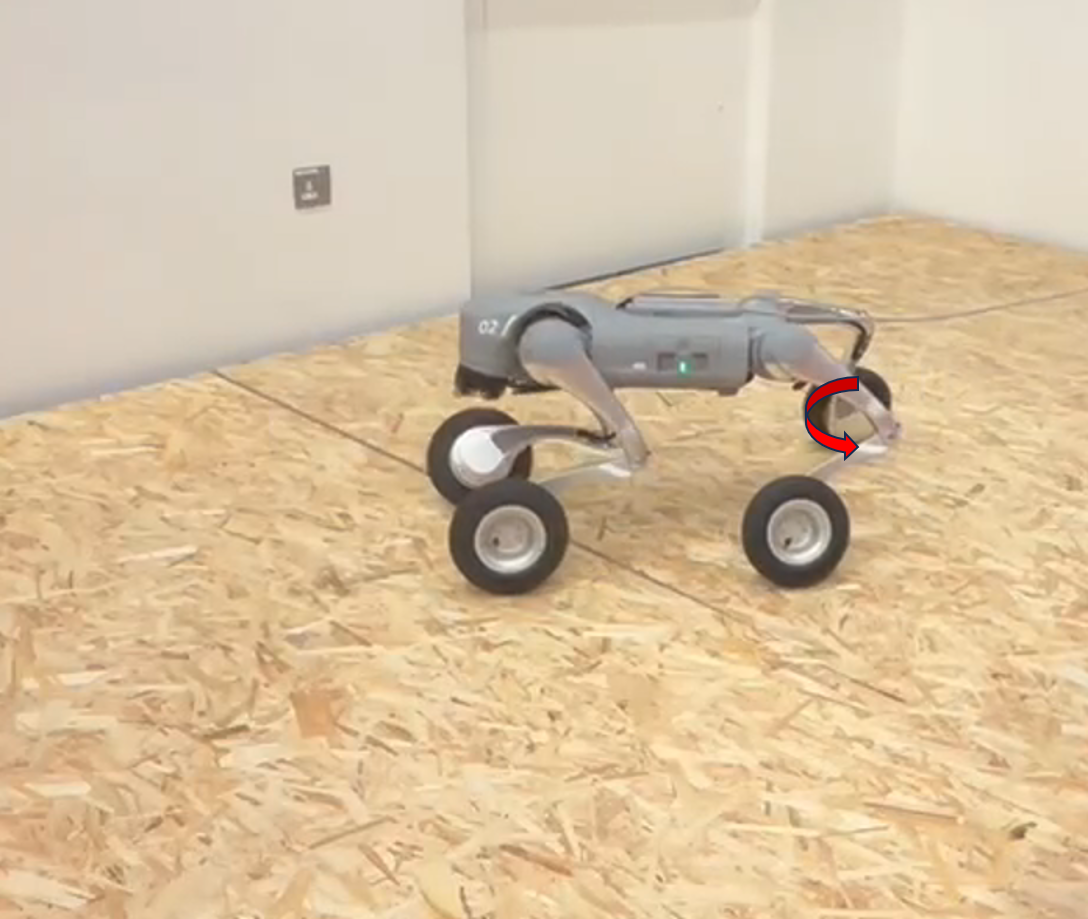}
        \caption{}
    \end{subfigure}
    % \hspace{1pt}
    \begin{subfigure}[b]{0.15\textwidth}
        \centering
        \includegraphics[width=\textwidth]{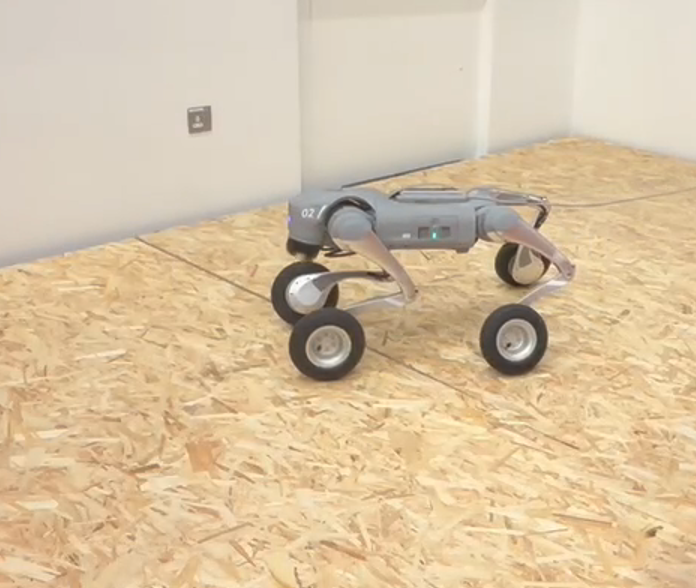}
        \caption{}
    \end{subfigure}
    \caption{Wheel and joint coordination for joint reset and Center-of-Mass adjustment during the recovery process.}
    \label{fig:kyon_recovery}
\end{figure}
In this paper, we explore a widely used learning-based approach to facilitate rapid post-fall recovery for wheeled-legged robots by leveraging their wheels. While most existing approaches struggle with sparse rewards, unstable convergence, and poor transferability across platforms, our method introduces an asymmetric Proximal Policy Optimization (PPO) framework that integrates dynamic episode exploration and curriculum learning. This design mitigates the sensitivity of recovery and smoothness weight parameters while also avoiding an overly conservative policy tendency. The main contributions of this work are summarized as follows:

\begin{itemize}
\item Adaptive Learning Framework: We propose an Episode-based Dynamic Reward Shaping mechanism combined with curriculum learning to balance exploration and policy convergence. This framework facilitates broad discovery of diverse recovery maneuvers in early training and gradually prioritizes precise posture control as learning progresses. It effectively mitigates the pitfalls of sparse reward formulations and unstable convergence common in prior reinforcement learning (RL)-based recovery methods.
\item Quantified Wheel–Leg Synergy: We systematically validate the critical role of wheel actuation in reducing joint torque effort and improving stabilization during recovery. This synergy between wheel dynamics and leg adjustments is rigorously quantified through cross-platform experiments, offering new insights into how wheels can actively assist in upright recovery processes.
\item Cross-Platform Generalization: We conduct extensive experimental validation on two distinct wheeled-legged quadrupeds (KYON and Unitree Go2-W\cite{Unitree})  demonstrating high adaptability to diverse hardware configurations. The proposed framework achieves recovery success rates above $97\%$ and substantial torque reduction without platform-specific tuning, underscoring its robustness, scalability, and real-world deployability.

%Extensive experimental validation on two distinct wheeled-legged platforms (KYON and Unitree Go2-W\cite{Unitree}) demonstrates the high adaptability of the proposed approach to varying hardware configurations, achieving higher success rates and reduced torque usage without platform-specific tuning, underscoring its robustness and scalability.
\end{itemize}

\section{RELATED WORK}
The post-fall recovery process typically requires a robot to execute a series of actions, gradually transitioning from an initial fallen state to a final standing posture. This process can be formulated as minimizing the discrepancy between the current state and the target state (e.g., base height, joint angles).
Due to the highly nonlinear kinematics and dynamics of legged robots, as well as the nonconvex nature of the recovery task itself, some optimization-based methods have resorted to using predefined motion sequences\cite{semini2015design,castano2019design} or simplified models\cite{saranli2004model} to reduce the complexity of optimization while achieving dynamic responsiveness and stable balance. However, these heuristic approaches generally depend heavily on model accuracy and significant manual effort, and they often lack portability across different platforms or tasks.

In recent years, learning-based approaches have shown great promise as alternatives, owing to their model-free nature and the ability to directly learn policies through interaction with the environment. For instance, Lee et al.\cite{lee2019robust} employ a hierarchical mechanism to decompose the task into two controllers: one for self-righting and one for standing. Hwangbo et al.\cite{hwangbo2019learning} propose using a single controller to optimize the base height, supplemented with auxiliary rewards to generate more natural recovery motions. Their work also introduces curriculum learning and utilizes a network actuator in simulation to bridge the gap between simulated and real-world performance. More recently, \cite{ma2023learning}incorporated a manipulator into the recovery process and formulated it as a finite-horizon task, which not only improved recovery success rates but also reduced joint torque consumption. While these approaches focus on flat terrains, our subsequent experiments extend validation to include non-flat scenarios, addressing challenges from uneven ground conditions.

In reinforcement learning–based recovery, using only the terminal posture as a reward signal results in sparse and delayed feedback, which hinders exploration and limits policy robustness.
 To overcome this, we propose an Episode-based Dynamic Reward Shaping mechanism that adapts reward composition throughout training to balance exporation and convergence. This approach promotes broad discovery of recovery maneuvers in the early stages, while gradually and gradually emphasizes precise posture refinement ensuring that the policy effectively converges to the desired recovery behavior.
\begin{figure}[t]
    \centering
    \begin{subfigure}[b]{0.15\textwidth}
        \centering
        \includegraphics[width=\textwidth,height=0.8\textwidth,keepaspectratio=false]{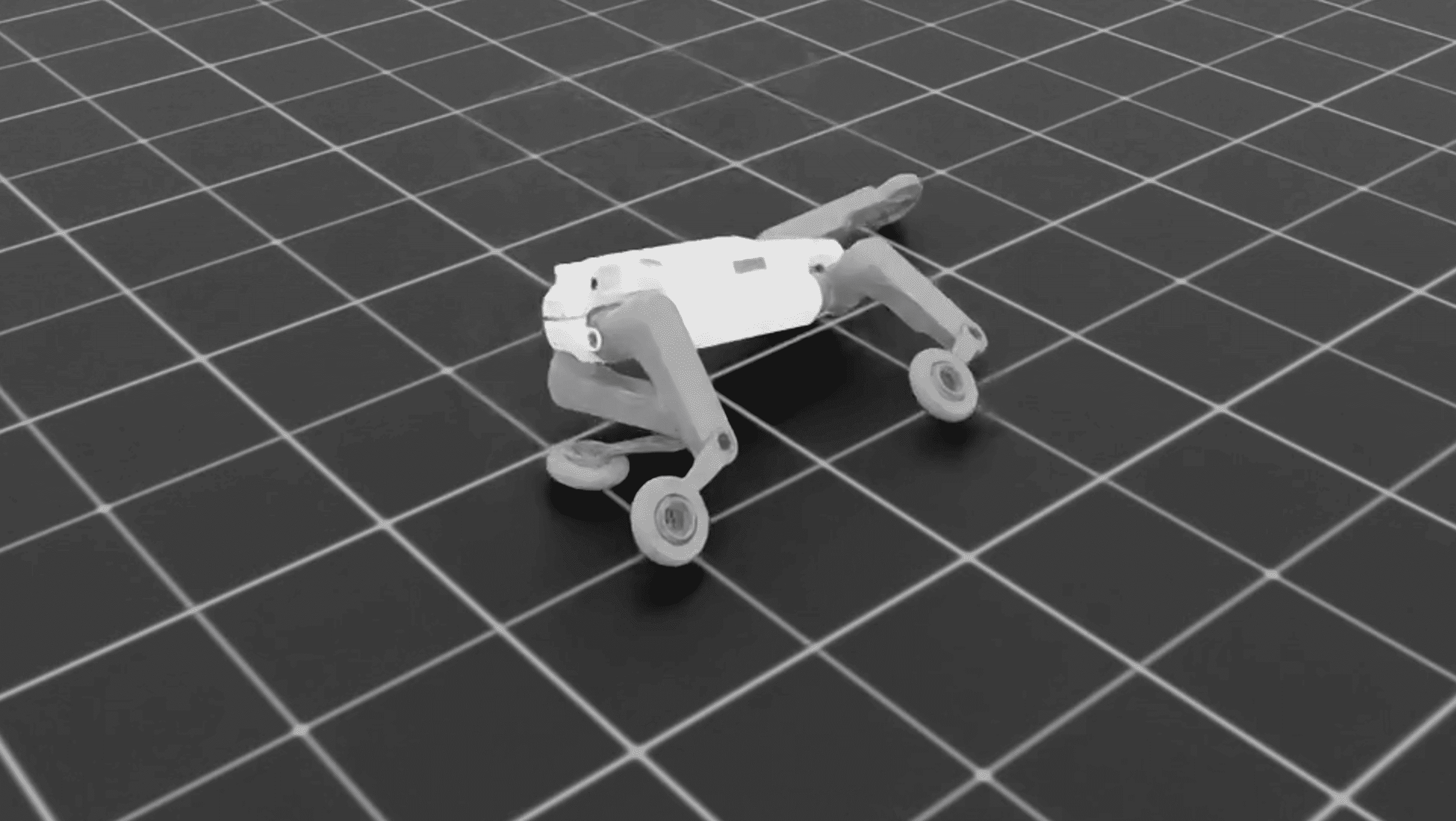}
        \caption{}
    \end{subfigure}%
    \hspace{-0.04cm}
    \begin{subfigure}[b]{0.15\textwidth}
        \centering
        \includegraphics[width=\textwidth,height=0.8\textwidth,keepaspectratio=false]{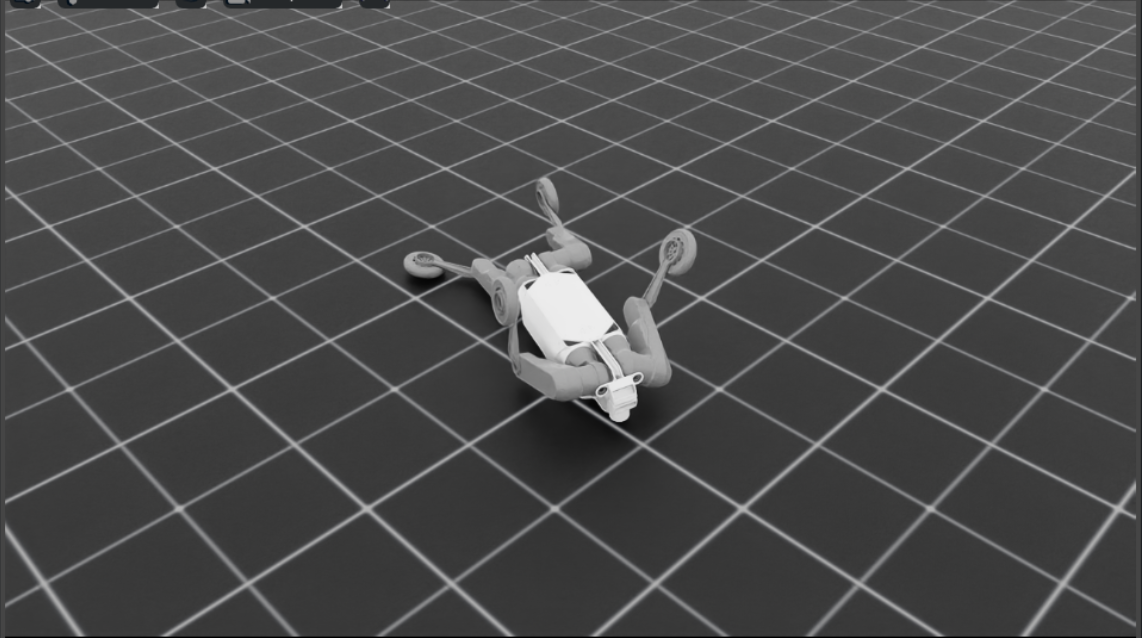}
        \caption{}
    \end{subfigure}%
    \hspace{-0.04cm}
    \begin{subfigure}[b]{0.15\textwidth}
        \centering
        \includegraphics[width=\textwidth,height=0.8\textwidth,keepaspectratio=false]{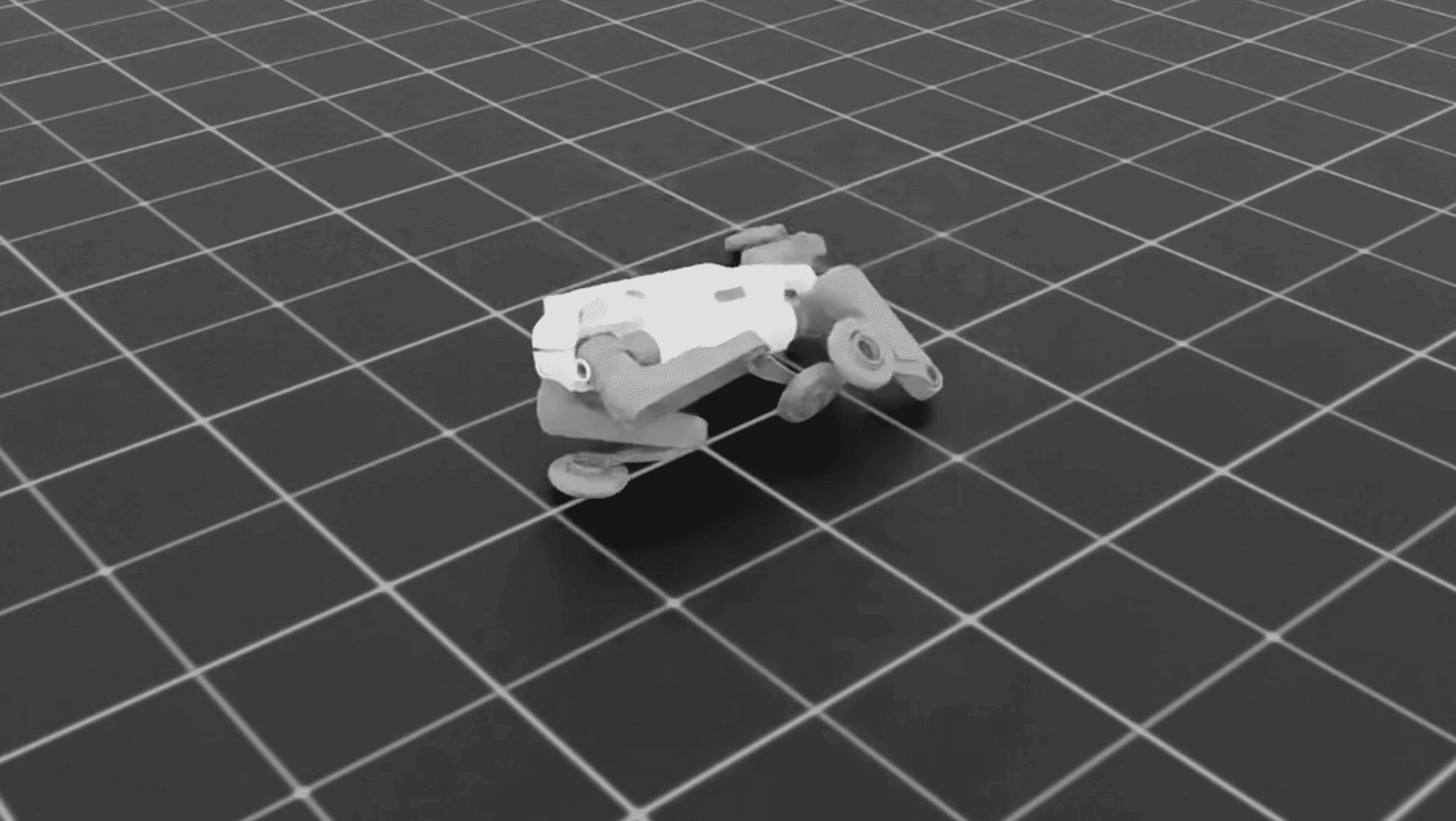}
        \caption{}
    \end{subfigure}
    \caption{(a)-(c) correspond to the initial states of different episodes. KYON Legged-wheeled Robot model: The presented model corresponds to a new robot under development.}
    %  in the Humanoid and Human Centered Mechatronics laboratory at Istituto Italiano di Tecnologia
    \label{initial_pos}
\end{figure}

\section{METHOD}
\begin{figure}[t]
    \centering
    \includegraphics[clip,width=0.48\textwidth]{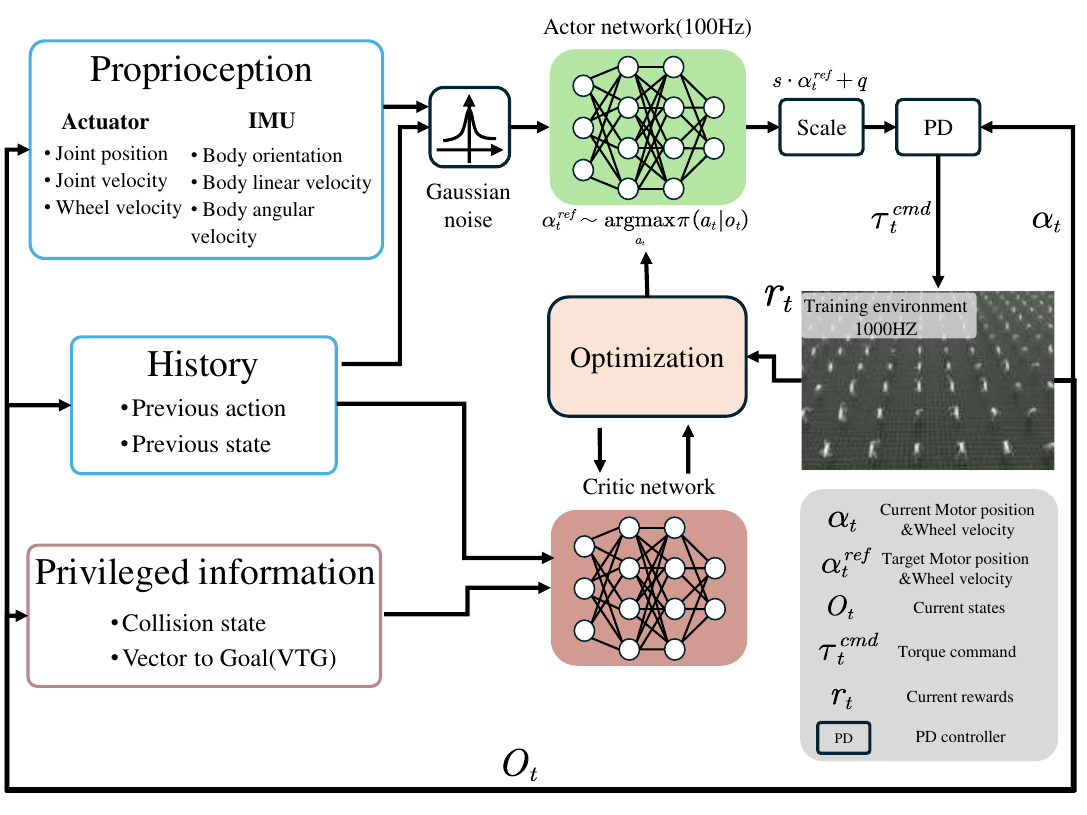}
    \caption{Asymmetric PPO-Based Reinforcement Learning Training Framework}
    \label{fig:frame}
\end{figure}

Fig.~\ref{fig:frame} provides an overview of our training pipeline. The implementation in this work is based on the Isaac Lab framework\cite{mittal2023orbit}. In the following, we outline the main components of the training process.
\subsection{State initialization and rollout} During the execution of various tasks, wheeled-legged robots may fall or even completely roll over; thus, the recovery controller must accommodate various initial conditions. To simulate different fallen states, we randomly initialize the robot’s base orientation and joint angles, set the joint torques to zero, and let the robot free-fall from a height of 1.1 m for 2 seconds. This duration is chosen to ensure that the robot fully collapses onto the ground, as shown in Fig.~\ref{initial_pos}, which helps avoid the situation where the dynamics become unrealistic\cite{hwangbo2019learning}. The diverse set of observed states improves policy generalization and adaptability. Each episode has a fixed duration of 5 seconds and is not terminated early.
\subsection{Privileged information based Actor-Critic}We employ a PPO framework featuring an asymmetric actor-critic architecture, where the critic network leverages privileged information to produce more accurate Q-value estimates and thereby expedite training. The observation inputs for the actor and critic are as follows:

\textit{Actor Observation}: The policy network observes the robot’s state, including current and historical readings (with a time interval of 0.01 s) of the base’s linear velocity, angular velocity, orientation, joint positions, joint velocities, and wheel speeds. This temporal information enables the robot to better evaluate its contact state\cite{hwangbo2019learning}. The actor also receives the previous action taken. Except for the most recent action and joint positions, Gaussian noise is added to the observations to account for state-estimation inaccuracies that arise after a fall. Similarly to\cite{christensen2021roadmap}, we apply higher noise levels specifically to joint velocities and the base’s linear and angular velocities.

\textit{Critic Observation}: The privileged information comprises data accessible only in simulation (and not during real-world deployment). By using this information, the critic can estimate Q-values more accurately in the training phase, thereby guiding the actor’s actions more effectively. The privileged observations include collision states, which to avoid unrealistic actions that may lead to damaging ground impacts, and VTG information\cite{xiao2020thinking} to accelerate learning.
\subsection{Actions}The target values for the joint positions and wheel velocities are computed as $s_p\cdot \mathbf{a_p}+\mathbf{q}$ and $s_v\cdot \mathbf{a_v}+\mathbf{q_v}$, respectively. Here $s_p$ and $s_v$ are scaling factors, $\mathbf{a}$ denotes the action with the highest probability from the policy network’s output distribution, and $\mathbf{q}$ and $\mathbf{q_v}$ represent the default joint angles and default wheel speed. The computed joint position commands are sent to the driver’s position PD controller, while the wheel velocity commands are transmitted to the motor speed controller.

In our experiments, we observed that when $s_v>1.0$, the action values become highly sensitive to changes in velocity control, resulting in noticeable wheel oscillations. Therefore, we chose $s_v<1.0$ to reduce sensitivity to command variations and enhance motion stability.
\subsection{Dynamic Exploration strategy}Building on an existing reward structure, this study introduces a dynamic episode factor to enhance exploration, thereby extending previous work\cite{christensen2021roadmap,ma2023learning,hwangbo2019learning,xiao2020thinking}. Compared with traditional sparse rewards, our approach allows greater deviation in posture during the early stages so that the robot can explore a variety of flipping or rolling maneuvers; it then progressively imposes tighter constraints on the target pose in later stages to guide the policy toward a robust fall-recovery solution. Empirical results show that this method trains a more resilient policy, as demonstrated in the ablation studies in Sec.\ref{rescompare}.

1) \textit{Dynamic Reward Shaping}:In previous research, the post-fall recovery process has often been simplified to a “final pose pursuit” task, typically employing relatively fixed reward or penalty terms based on joint angles, body height, or orientation error. However, in transitioning from a prone position to standing, a wheeled-legged robot may need to undergo large—and sometimes counterintuitive—rolling or swinging movements. Penalizing only the terminal pose too strongly can restrict exploration of diverse state-action pairs, as illustrated in Fig. 2. To balance free exploration in the early stage with precise posture control in the later stage, we propose Episode-based Dynamic Reward Shaping(ED), formally defined as:
\begin{equation}
    ed=\left ( \frac{a\cdot t}{T}  \right ) ^k, t\in [0,T]
    \label{eq1}
\end{equation}
where $t$ is the current step in an episode. $T$ is the total number of steps per episode, and $k$ serves as the growth rate, whereas $a$ is the baseline coefficient. For our experiments, we set $a=\frac{T}{2}$, according to the expected recovery time, $k=3$. The immediate reward at each time step is given by:
\begin{equation}
    r_t=ed\cdot R\left ( s_t,a_t \right ) 
    \label{eq2}
\end{equation}
Here, $R(s_t, a_t)$ is the original task-related reward function. When the $t$ is close to 0, the agent can attempt substantial posture adjustments, and any successful state-action pairs are retained and reinforced in the policy network. As $t$ approaches the maximum length of the episode, the emphasis is progressively increased on the refinement of the final target pose. This incremental mechanism promotes action diversity in the early stage of each episode and ensures the stability of the standing posture in the later stage.

We observe that this dynamic reward modulation can slow down learning efficiency. However, as discussed in Sec. III-B, empirical results indicate that it enhances the robustness of the learned policy. This effect can be partly attributed to broader coverage of the state space\cite{zhang2022accessibility}, meaning that more state-action combinations are explored during training, thereby improving the policy’s generalization capability for various fall scenarios.

2)\textit{Curriculum Learning}: To further improve training efficiency and mitigate the reduction in learning speed caused by the dynamic episode factor, we introduce curriculum learning\cite{hwangbo2019learning,lee2020learning,rudin2022learning}. The curriculum weight (CW) is defined as:
\begin{equation}
    cw=cw^{\beta} 
    \label{eq3}
\end{equation}
where $\beta$ is the progress rate. We multiply the behavior reward by this weight and update it before each environment reset. This approach ensures rapid acquisition of task-related behaviors in the early stages and smoother motion in the later stages, striking a balance between efficiency and feasibility.
\begin{table}[t] % 表格在单列模式中
\centering
\renewcommand{\arraystretch}{1.3} % 调整行间距
\setlength{\tabcolsep}{4pt} % 调整列间距
\caption{Reward Terms Summary}
\label{tab:reward_terms}
\begin{tabular}{@{}ccc@{}} % 列宽调整
\textbf{Reward Term} & \textbf{Definition} & \textbf{Scale} \\ 
\midrule
stand joint position & 
$e^{-\frac{\sum_{j}(\mathbf{q}_j^* - \mathbf{q}_j[t])^2}{\sigma_{p}N_j}}$ 
& 
42 \\

base height &
$e^{-\frac{\max(h^* - h_b[t], 0)^2}{\sigma_{h}}}$ 
&
120 \\

base orientation &
$(\mathbf{g}_b -\mathbf{e}_z)^2$ &
50 \\

body collision &
$\sum_{b \in B} \|\lambda_b[t]\|^2$ & 
$-5 \times 10^{-2} $\\

action rate &
$\sum_{t}(\mathbf{a}[t] - \mathbf{a}[t-1])^2$ & 
$-1 \times 10^{-2}$ \\

joint velocity &
$\sum_{j} \dot{\mathbf{q}_j}^2$ & 
$-2 \times 10^{-2}$ \\

torques &
$\sum_{j} \tau_j^2$ & 
$-2.5 \times 10^{-5}$ \\

acceleration &
$\sum_{j} \ddot{\mathbf{q}_j}^2$ & 
$-2.5 \times 10^{-7}$ \\

wheel velocity &
$\sum_{k} \dot{\mathbf{q}_k}^2$ & 
$-2 \times 10^{-2}$ \\

\bottomrule
\end{tabular}
\end{table}
\begin{figure*}[!t]
    \centering
    \begin{minipage}[c]{0.2\textwidth}
        \centering
        \includegraphics[width=\textwidth]{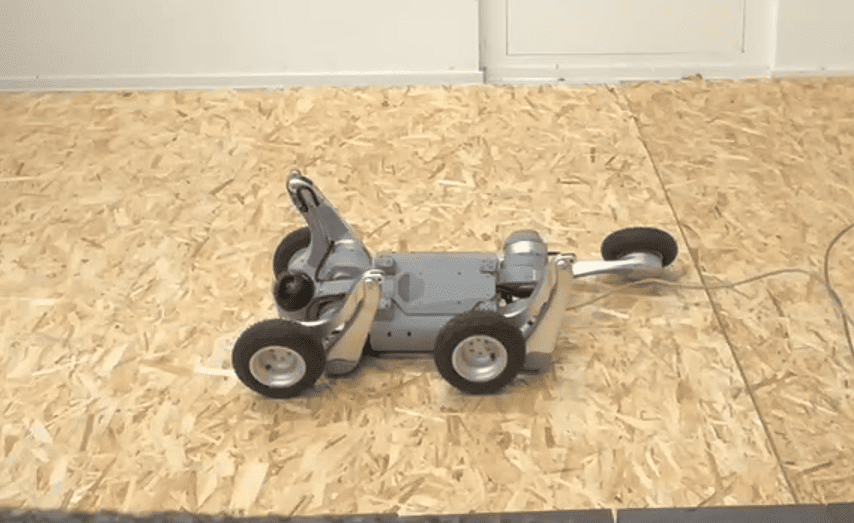}
        \label{compare:ini}
    \end{minipage}
    \hskip 0.1em
    \begin{minipage}[c]{0.7\textwidth} 
        \centering
        \begin{subfigure}[b]{0.23\textwidth}
            \centering
            \includegraphics[width=\textwidth]{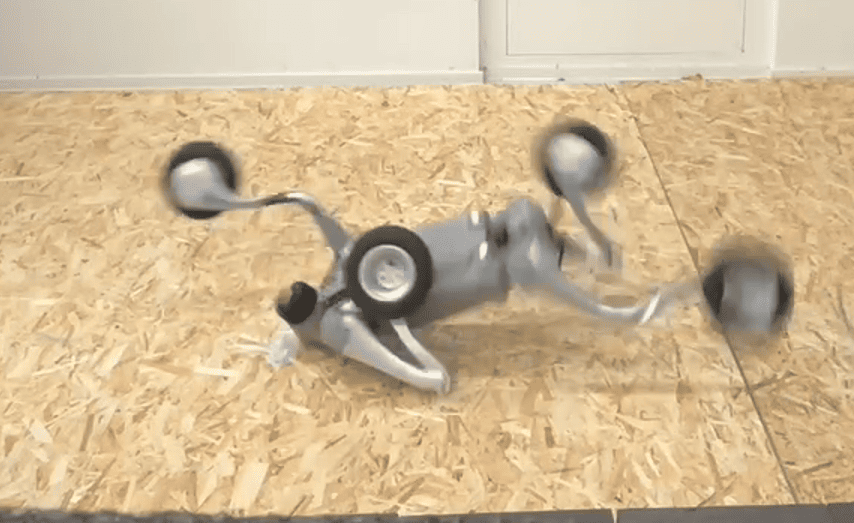}
            \caption{}
        \end{subfigure}
        \hspace{1pt}
        \begin{subfigure}[b]{0.23\textwidth}
            \centering
            \includegraphics[width=\textwidth]{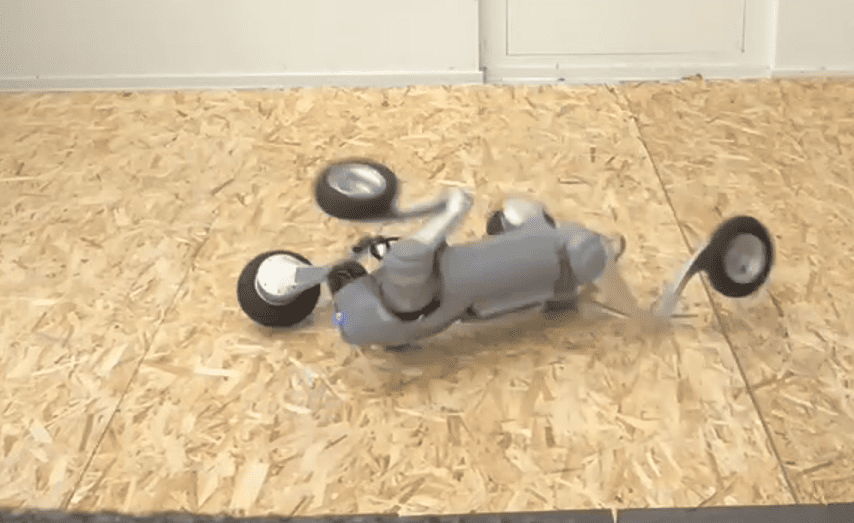}
            \caption{}
        \end{subfigure}
        \hspace{1pt}
        \begin{subfigure}[b]{0.23\textwidth}
            \centering
            \includegraphics[width=\textwidth]{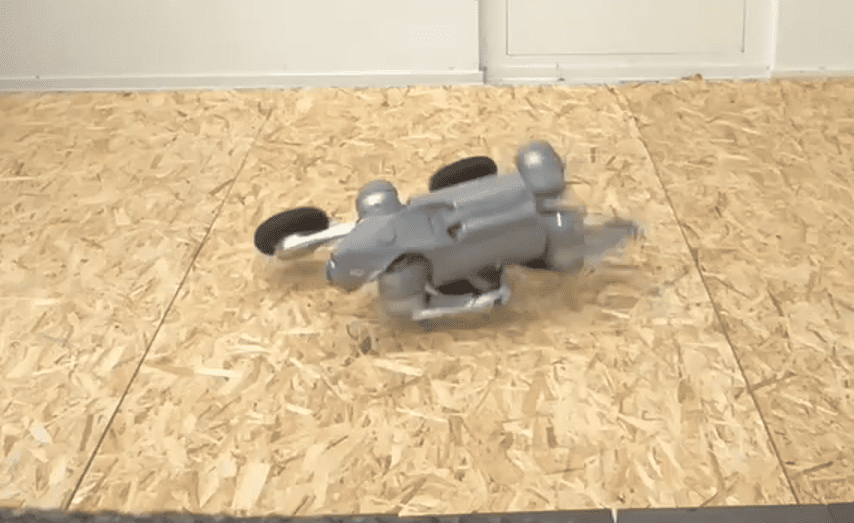}
            \caption{}
        \end{subfigure}
        \hspace{1pt}
        \begin{subfigure}[b]{0.23\textwidth}
            \centering
            \includegraphics[width=\textwidth]{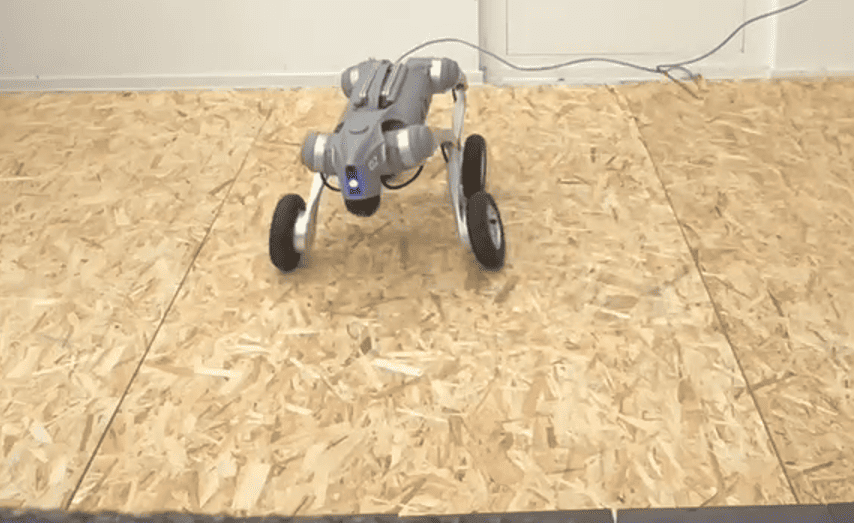}
            \caption{}
        \end{subfigure}
        \vskip 0.5em
        \begin{subfigure}[b]{0.23\textwidth}
            \centering
            \includegraphics[width=\textwidth]{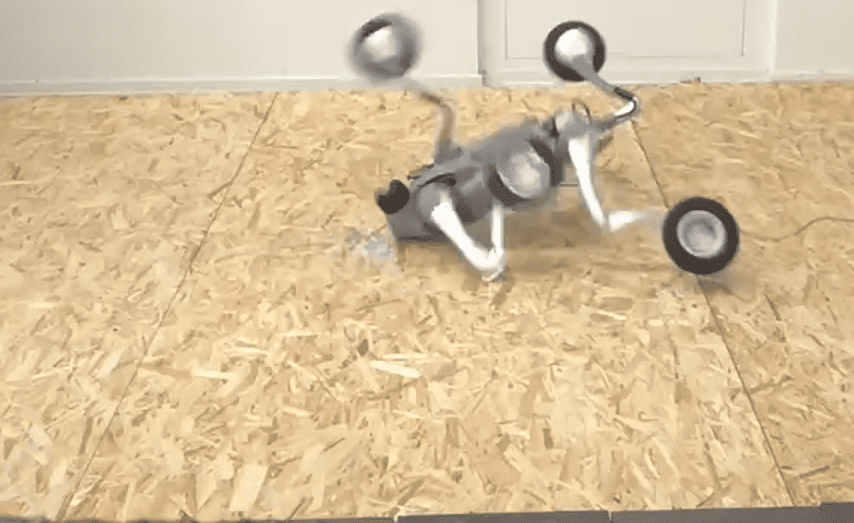}
            \caption{}
        \end{subfigure}
        \hspace{1pt}
        \begin{subfigure}[b]{0.23\textwidth}
            \centering
            \includegraphics[width=\textwidth]{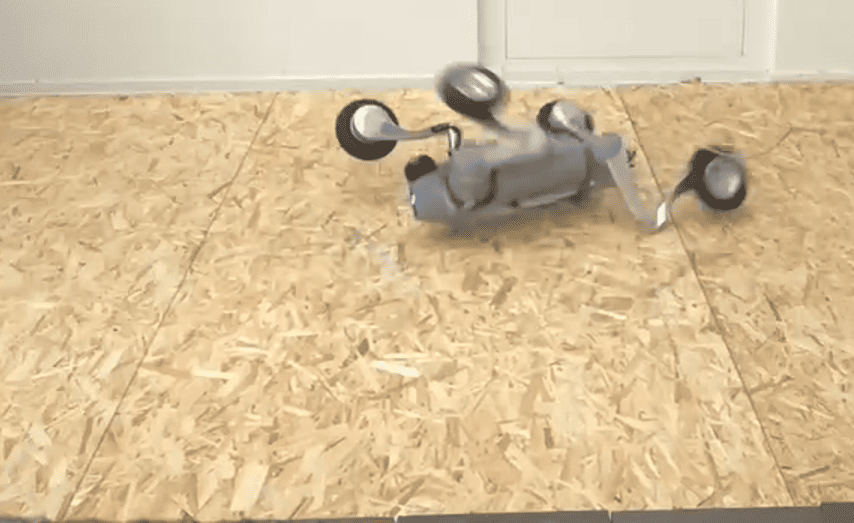}
            \caption{}
        \end{subfigure}
        \hspace{1pt}
        \begin{subfigure}[b]{0.23\textwidth}
            \centering
            \includegraphics[width=\textwidth]{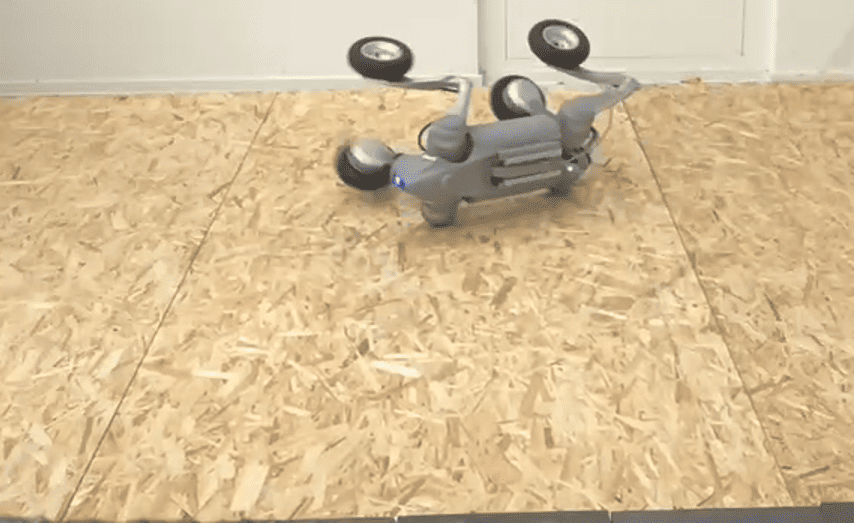}
            \caption{}
        \end{subfigure}
        \hspace{1pt}
        \begin{subfigure}[b]{0.23\textwidth}
            \centering
            \includegraphics[width=\textwidth]{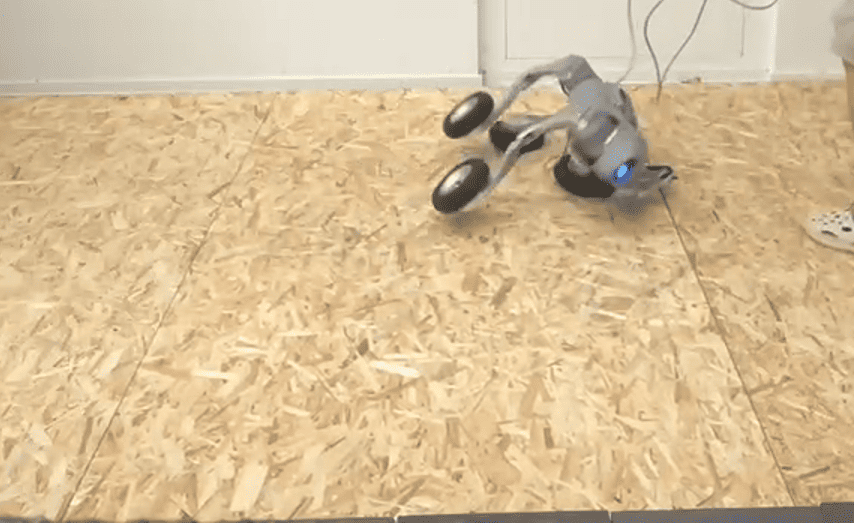}
            \caption{}
        \end{subfigure}
    \end{minipage}
    \caption{Recovery processes under the same initial posture using two different strategies. (a)–(d) shows the ED strategy successfully recovery. (e)–(h) illustrates the baseline strategy, which adjusts joint positions and base height but fails to optimize base orientation, becoming trapped in a local optimum and not fully recovering. Full videos could be found on the webpage.}
    \label{fig:compare}
\end{figure*}
\subsection{Reward function} We divide the reward into task rewards and behavior rewards, designed respectively to incentivize correct movements while constraining excessively aggressive behavior. All the robot-state variables required for these computations—such as body acceleration and contact forces—are directly obtained from the simulation environment. The rewards are categorized as follows:

1)\textit{Task Rewards}: To encourage the robot to quickly and accurately achieve the target posture, we define the following three primary sub-rewards:

\textit{Base Height}: Penalizes deviations between the base height and the desired height to ensure the robot stands upright.

\textit{Base Orientation}: Penalizes misalignment of the gravitational projection to correct roll and pitch angles.

\textit{Joint Position}: Penalizes deviations from the default joint angles.

2)\textit{Behavior Rewards}: To encourage smoother recovery, the behavior rewards penalize measures such as the action rate and joint acceleration. When computing the action rate, the wheel actions are excluded to allow the robot to have greater flexibility in using its wheels for recovery. We also introduce a penalty for body collisions to prevent severe impacts with the ground or self-contact that could damage the robot’s structure. Additionally, to stabilize the robot once upright, we provide a reward for the support state, defined as the condition where all four wheels are in contact with the ground simultaneously.

Owing to the introduction of the dynamic exploration mechanism, during the early training stages, the agent initially focuses on utilizing joint motions to reorient the body at the beginning of each episode, and gradually shifts its attention toward refining joint angles and base height in the later phase of the episode.

Table \ref{tab:reward_terms} summarizes all reward terms and their scaling. In the reward definitions, $\mathbf{q}_j$, $\dot{\mathbf{q}_j} $ and $\ddot{\mathbf{q}_j} $denote the angular position, velocity, and acceleration of each joint. $\mathbf{g}_b$ is the gravity vector projected onto the base frame of the robot. $B$ denotes the set of selected robot links, including the shanks, thighs, and base. $\lambda_b$ represents contact force for body $b$.

Finally, the overall reward at each time step is defined as:
\begin{align}
    r_t = & \, ed \cdot( r_{joint\_pos} + r_{base\_height} + r_{base\_ori}) \notag\\
          & + cw \cdot \big( r_{collision} + r_{action\_rate}\notag\\
          &  + r_{joint\_vel} + r_{torque} + r_{joint\_acc}) \notag\\
          & + r_{wheel vel}
    \label{eq4}
\end{align}

\section{RESULTS}
\subsection{Experimental Setup}
\subsubsection{Simulation Setup}
All experiments are conducted in the Isaac Sim simulator using KYON and Unitree Go2-W\cite{Unitree} robot models. To enhance policy robustness, extensive domain randomization is applied: the robot’s base mass is perturbed by $\pm5.0 kg$ at the start of each episode, link masses vary by $\pm10\%$ from their nominal values. Moreover, contact dynamics are randomized by uniformly sampling static friction coefficients from $[0.7, 1.3]$ for all robot bodies, while actuator stiffness and damping are perturbed by $\pm1.0$ and $\pm0.1$ respectively. Additionally, action noise characterized by a zero mean and a standard deviation of $0.02$ is incorporated.

Each episode lasts 5 seconds (at a control rate of 50Hz) and terminates solely upon timeout. The training framework employs an asymmetric PPO algorithm with a learning rate of $0.001$ and a discount factor of $\gamma =0.99$. Furthermore, curriculum learning is integrated, beginning with a difficulty factor of $0.3$ that decays exponentially at a rate of $0.968$ to progressively enhance policy exploration.
\subsubsection{Sim-to-Real}
We deploy and evaluate the learned policy on a Unitree Go2W\cite{Unitree}, performing zero-shot sim-to-real transfer. The trained policy is available on the project website.
\subsection{Ablation of the Episode-based Dynamic Reward Shaping Mechanism}
\label{rescompare}
To verify the effectiveness of the ED, we compared the recovery performance of the ED-Policy with that of a baseline under an identical reward function and hyperparameter framework. Recovery was defined as achieving, at the final time step of an episode, a base height exceeding $0.42m$, a deviation of the joint angle from the default configuration of no more than $0.5rad$, and a maximum joint velocity below $0.1$ rad/s. Testing on $2048$ randomly generated initial conditions, the ED-Policy achieved a success rate of $99.1\%$, representing a $2.7$ percentage point improvement over baseline $96.4\%$.

Further analysis revealed that the failure cases for the baseline were predominantly associated with scenarios in which the front leg joint angles were minimally randomized, shown in Fig.~\ref{fig:compare}. To illustrate this, we selected a representative initial state from one of the baseline’s failure cases and used it as the starting state for the ED-Policy to compare the resulting action sequences. The experiments indicate that, due to the penalty imposed on joint angle deviations, the baseline suppressed front leg motions and attempted to adjust the base height solely through the hind legs, ultimately becoming trapped in a local optimum.

\begin{figure}[ht]
    \centering
    \includegraphics[width=0.43\textwidth]{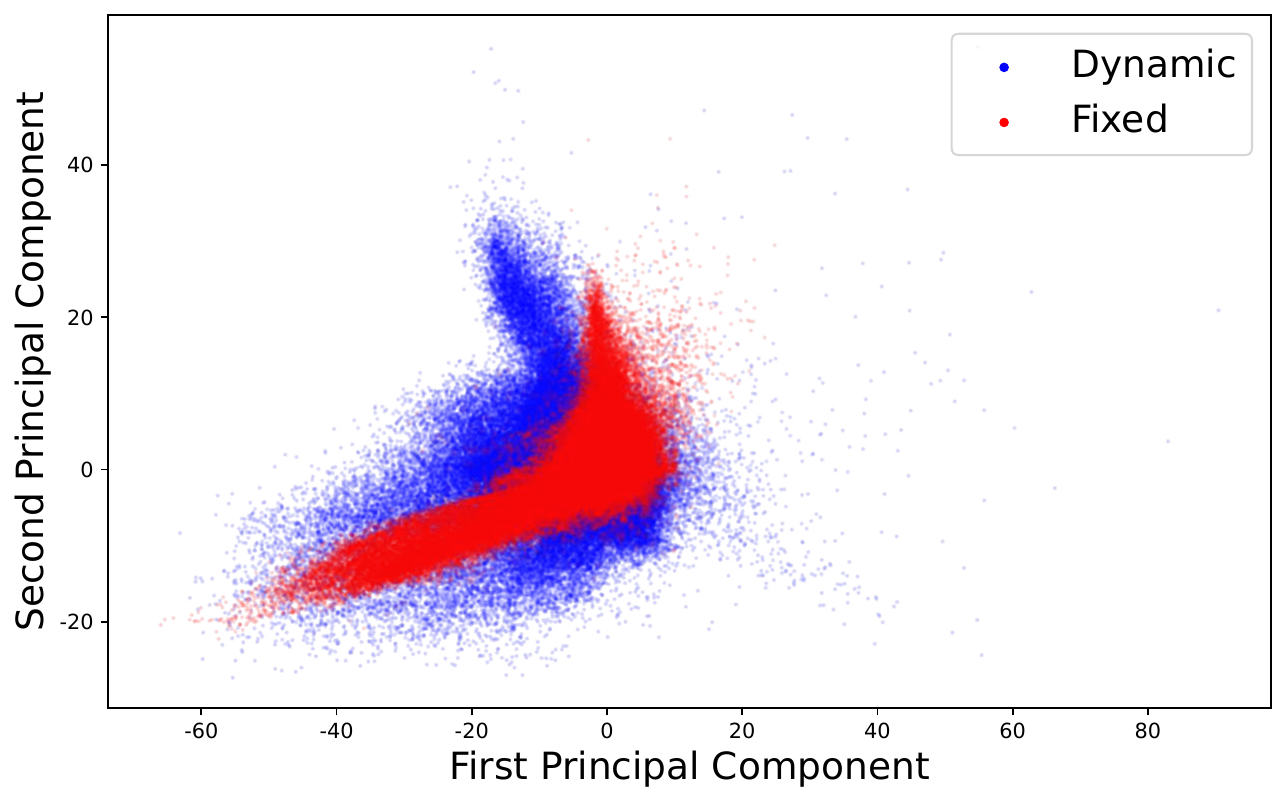} 
    \caption{PCA-based comparison of single-episode action distributions for ED-policy and baseline-policy across 2048 environments.}
    \label{fig:pca}
\end{figure}

This phenomenon is directly related to differences in the exploration capacity of the action space. Principal Component Analysis (PCA) was applied to the action sequences from the 2048 experiments, and the resulting visualization Fig.~\ref{fig:pca} shows that the variance along the principal components for the ED-Policy ($\sigma ^2_{PC1}=20.91$, $\sigma ^2_{PC2}=9.72$) is significantly higher than that for the baseline ($\sigma ^2_{PC1}=17.33$, $\sigma ^2_{PC2}=5.22$). These findings confirm that the dynamic reward shaping mechanism, by expanding the exploration boundaries and generating less conservative action sequences, enhances recovery robustness in complex scenarios.
\subsection{Wheel-Assisted Recovery}
\begin{figure}[ht]
    \centering
    \includegraphics[width=0.43\textwidth]{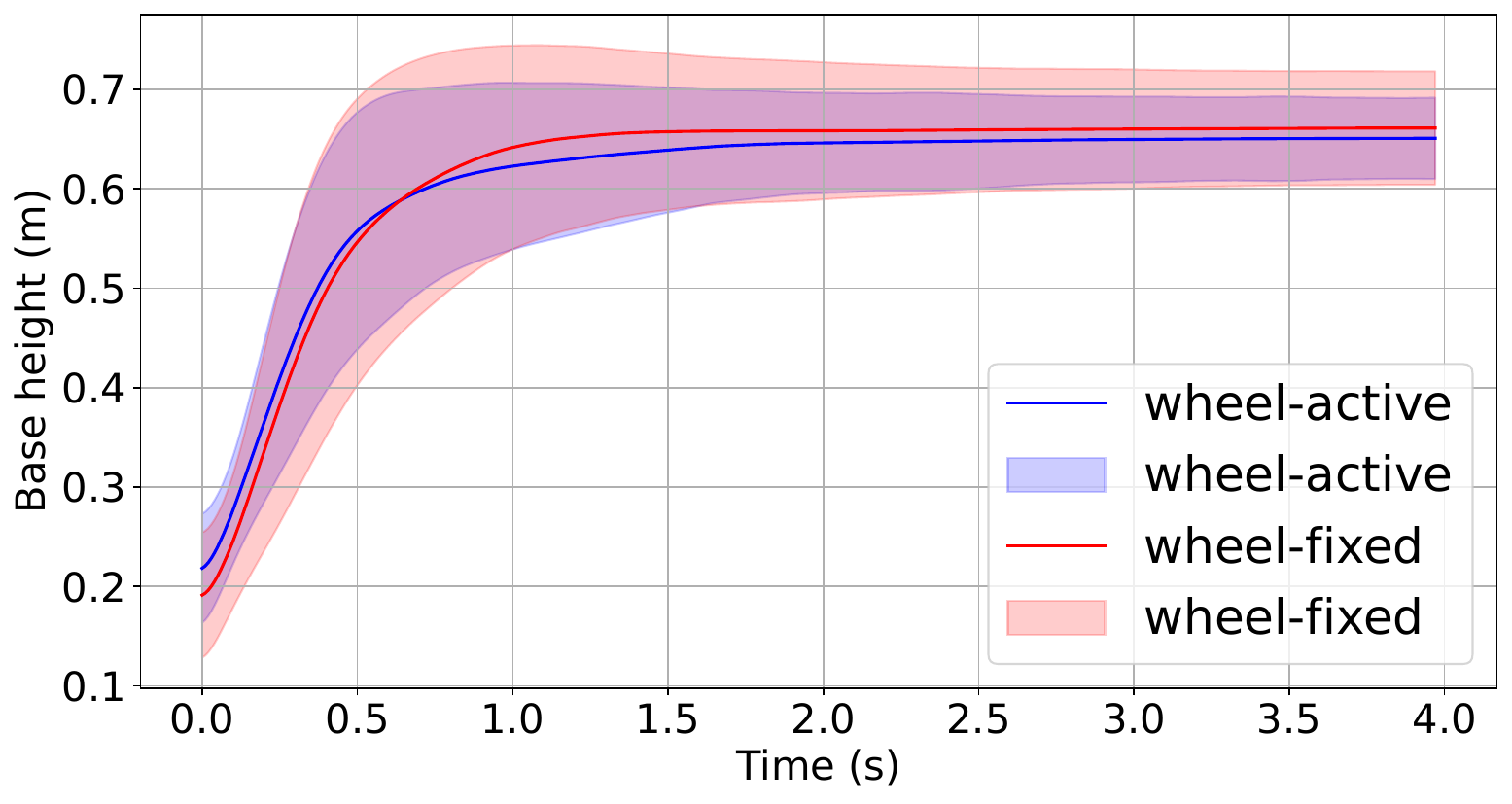}
    \caption{Comparison of base height dynamics, quantified by mean and variance, for pure legged and wheel–leg coordinated modes over a single episode across 2048 environments.}
    \label{fig:height}
\end{figure}

To evaluate the dynamic contribution of wheeled actuation during recovery, we compared the performance differences between a wheel–leg collaborative mode and a pure legged mode. In the experimental setup, the pure legged group was configured by modifying the robot description file to change the wheel joints from continuously actuated to fixed constraints, and by removing the wheel speed penalty term to eliminate policy bias; all other hyperparameters and reward functions were kept identical to those of the wheel–leg collaborative group. Testing across 2048 random initial states demonstrated that the wheel–leg collaborative mode significantly enhanced recovery stability and optimized joint torque distribution.
As shown in Fig.~\ref{fig:height}, although both strategies’ base height trajectories converged to the target height ($0.42 m$) within $0.7s$, the wheel–leg collaborative group exhibited lower variance during both the recovery phase ($t<0.7s$) and the stabilization phase ($0.7s<t<1.5s$). In contrast, the pure legged group showed larger fluctuations during the stabilization phase, with a variance of $0.102$ at $t=1s$. This indicates that the base height recovery process was considerably more stable, demonstrating that wheel-assisted rolling effectively mitigates the body oscillations observed in the pure legged recovery.

\begin{figure}[ht]
    \centering
    \includegraphics[width=0.43\textwidth]{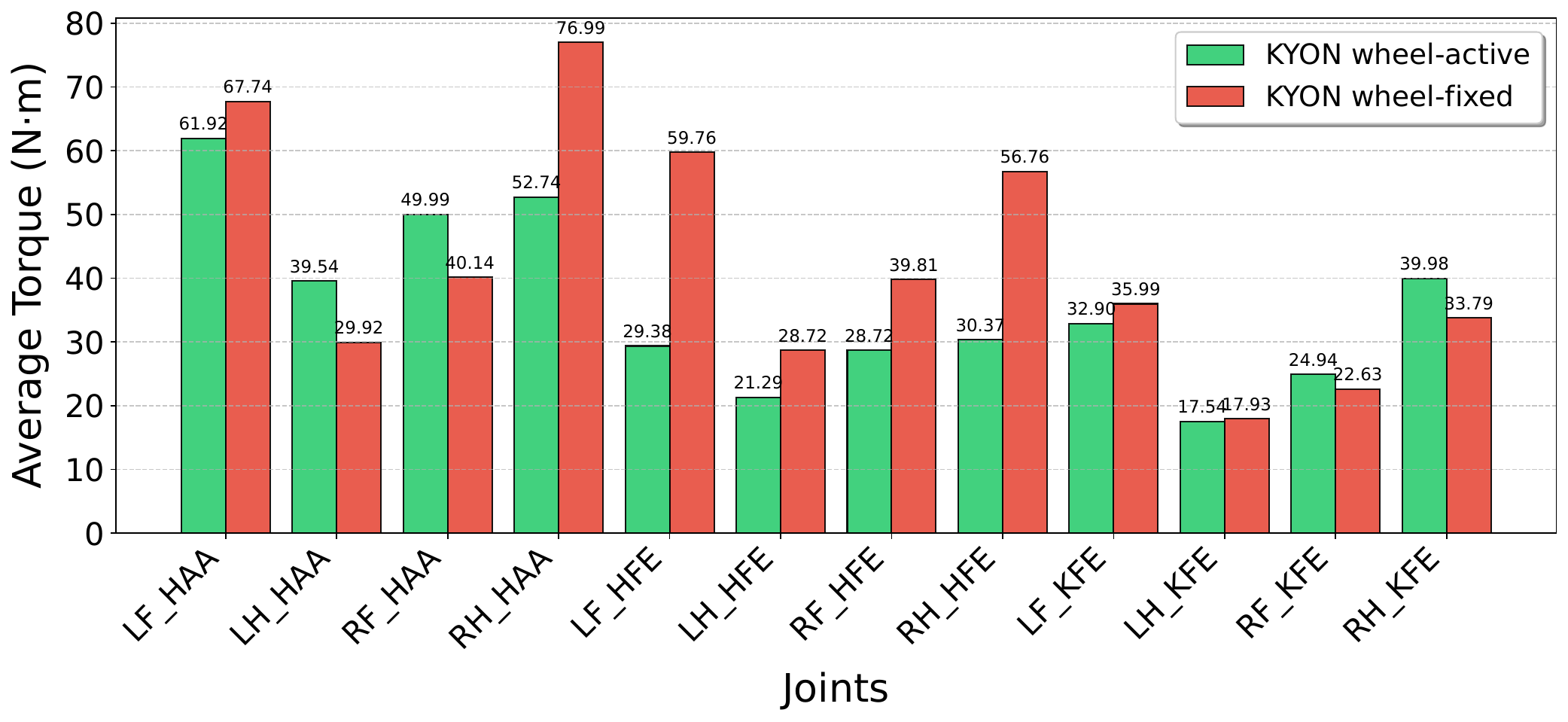}
    \caption{Comparison of average joint torque for KYON under pure legged and wheel–leg coordinated modes across 2048 environments in a single episode.}
    \label{fig:torque}
\end{figure}

Further analysis of joint torque characteristics shown in Fig.~\ref{fig:torque} revealed that the average joint torque in the wheel–leg collaborative group was $35.776N\cdot m$, representing a $15.85\%$ reduction compared to the pure legged group ($42.515N\cdot m$). This reduction may be attributed to the conversion of wheel kinetic energy into joint potential energy via wheeled rolling, which decreases the energy consumption required by the joints to counteract gravity. It may be also due to the fact that the active wheels permit the sliding of the leg contacts benefiting the reduction of the internal forces, which can be larger and additionally stress the leg joints when the leg contacts cannot slide in the case of the fixed wheels,  thereby validating the role of wheel assistance in recovery control.
\subsection{Cross-Platform Validation on Different Robots}
\begin{figure}[ht]
    \centering
    \includegraphics[width=0.43\textwidth]{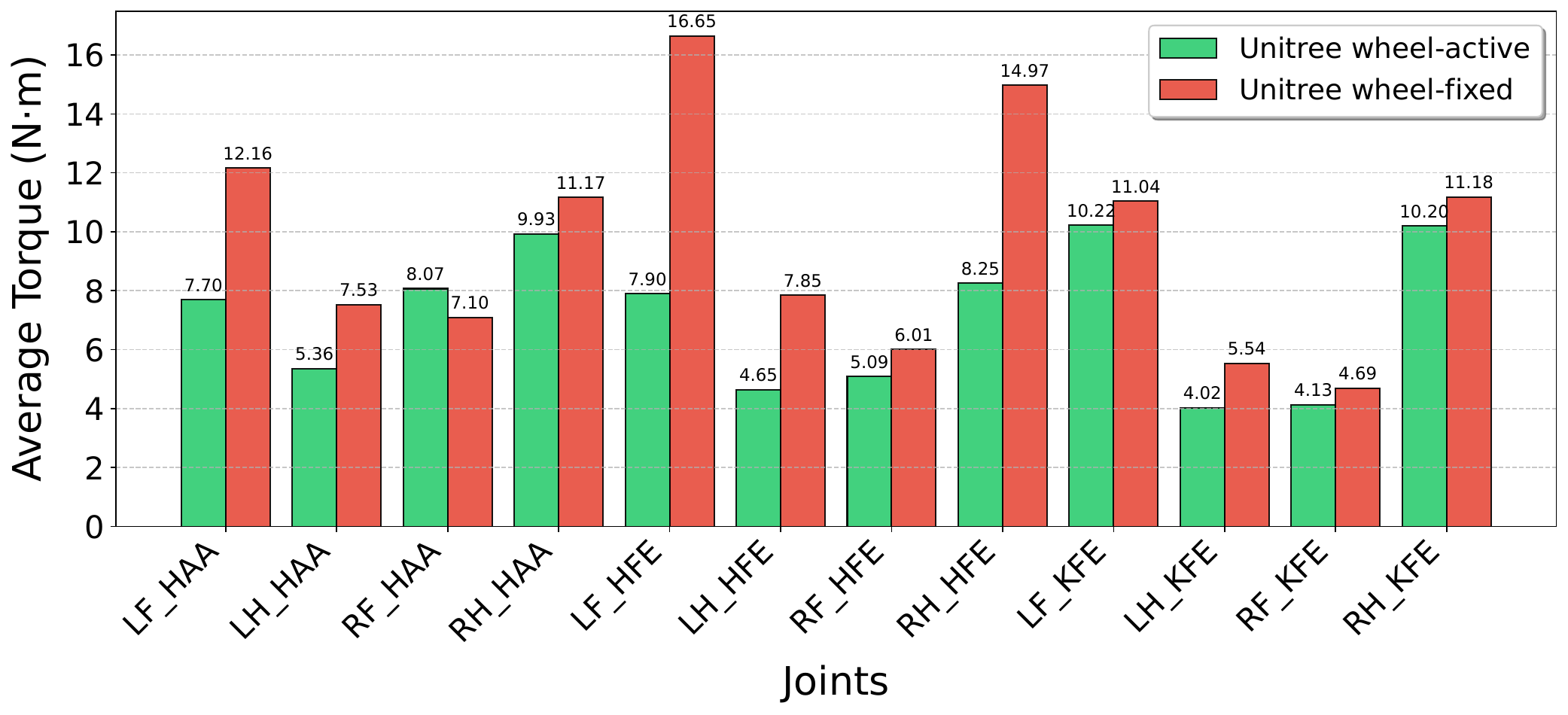}
    \caption{Comparison of average joint torque for Unitree Go2-W\cite{Unitree} under pure legged and wheel–leg coordinated modes across 2048 environments in a single episode.}
    \label{fig:torqueunitree}
\end{figure}

To validate the cross-platform generalization capability of the ED strategy, we deployed it on the Unitree Go2-W\cite{Unitree} platform with significantly different configuration parameters and compared its performance to a baseline in which the ED modules were ablated to isolate the effect of our proposed strategy. The experimental results indicate that despite hardware differences such as mass distribution and joint drive limits, the ED strategy maintained robust recovery performance and demonstrated the advantages of the wheel–leg collaboration mechanism. As illustrated in Fig.~\ref{unitree}, with the wheels actuated, the Go2-W robot utilized active wheel rotation to buffer its posture, thereby significantly reducing the magnitude of torque adjustments required at the hip joints. Its recovery success rate improved by $3.7$ percentage points compared to the baseline strategy, see Table \ref{table suc rate}, and the average joint torque was reduced by $26.2\%$ relative to the fixed-wheel configuration (see Table \ref{table torque com} and Fig.~\ref{fig:torqueunitree}). 

These findings are consistent with the experimental results on the KYON platform, indicating that the ED strategy consistently generates optimized action sequences tailored to local degrees of freedom via dynamic reward shaping, and that the wheeled assistance contact force regulation mechanism effectively reduces joint loads.
\begin{figure}[ht]
    \centering
    \includegraphics[width=0.43\textwidth]{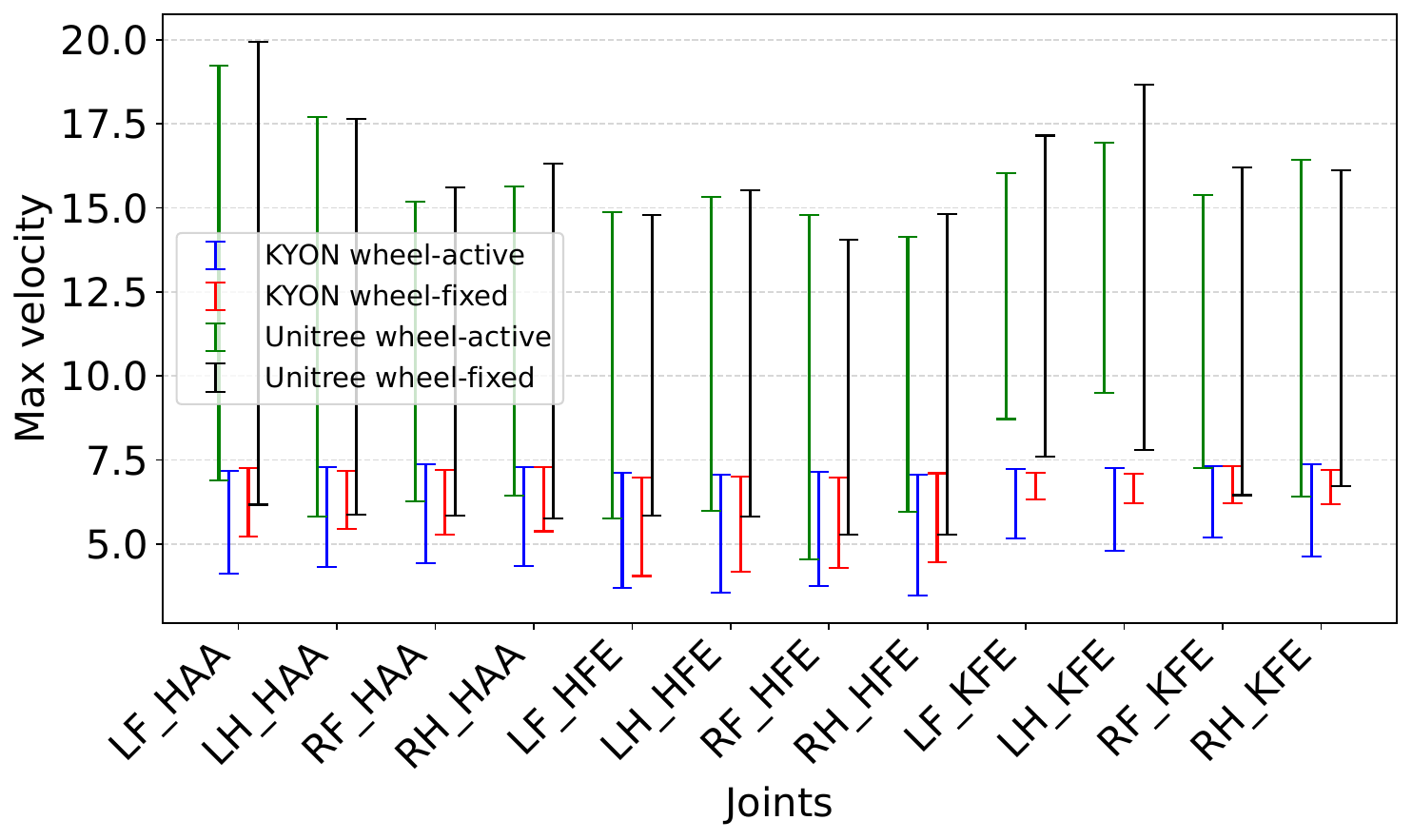}
    \caption{Mean $\pm$ standard deviation of maximum joint velocities for two robots in different driving mode across 2048 environments.}
    \label{fig:maxvel}
\end{figure}
\begin{figure}[ht]
    \centering
    \includegraphics[width=0.43\textwidth]{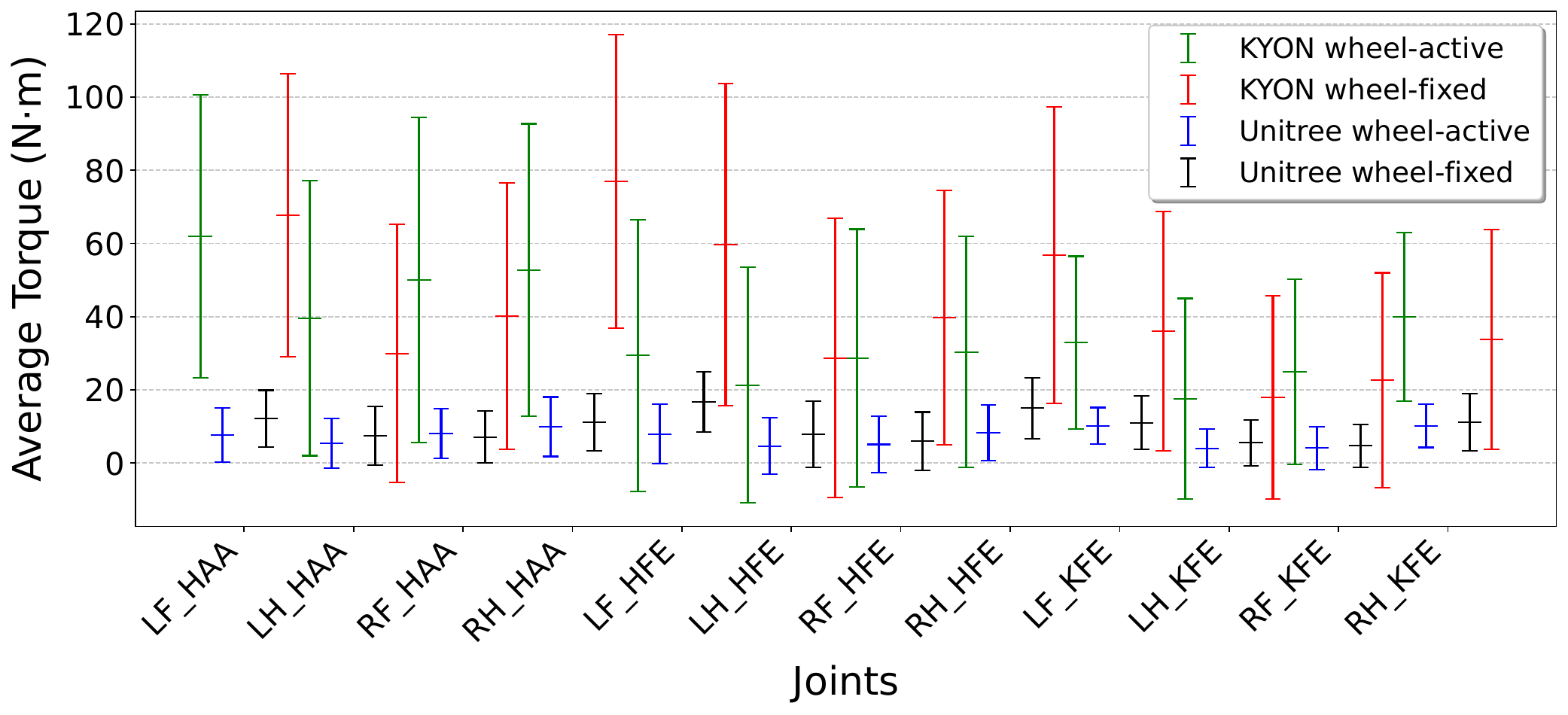}
    \caption{Mean $\pm$ standard deviation of joint torque for two robots in different driving mode across 2048 environments.}
    \label{fig:meantorque}
\end{figure}
We also performed a quantitative analysis of the joint motion parameters for both robots. As shown in Fig.~\ref{fig:maxvel}, by calculating the mean and standard deviation of the peak velocities for each joint, we verified that the driving commands generated by our control strategy adhered to the maximum velocity thresholds specified in the robot description files (KYON: $7.6 rad/s$ and Unitree Go2-W: $20.3 rad/s$). Furthermore, we also evaluated the mean and standard deviation of the torques for each joint, as illustrated in Fig.~\ref{fig:meantorque}. For the Unitree Go2-W robot, most joints exhibit a maximum torque of $23.7 N\cdot m$,  with the calf joint(KFE) reaching up to $35.55 N\cdot m$. In contrast, all joints of the KYON robot achieves a maximum joint torque of $185 N\cdot m$. All movements stay within the allowable joint range limits, confirming that the actions generated by our simulation-based control policy are indeed feasible for real-world experiments.
\begin{figure}[t]
    \centering
    \begin{subfigure}[b]{0.15\textwidth}
        \centering
      \includegraphics[trim=4cm 1 4cm 1,clip,width=\textwidth,height=0.8\textwidth,keepaspectratio=false]{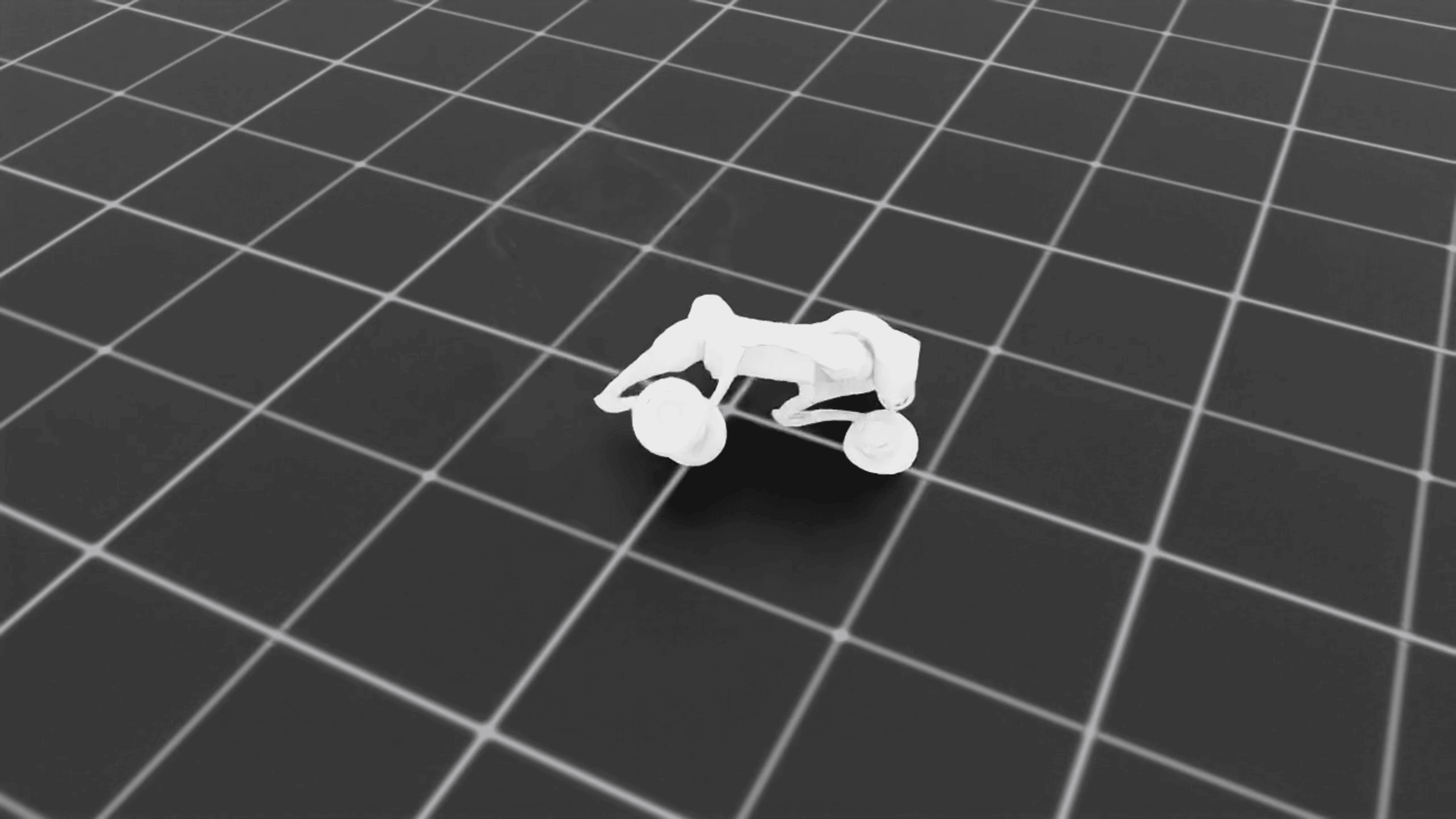}
        \caption{}
    \end{subfigure}%
    \hspace{0.1cm}
    \begin{subfigure}[b]{0.15\textwidth}
        \centering
        \includegraphics[trim=5cm 1 3cm 1,clip,width=\textwidth,height=0.8\textwidth,keepaspectratio=false]{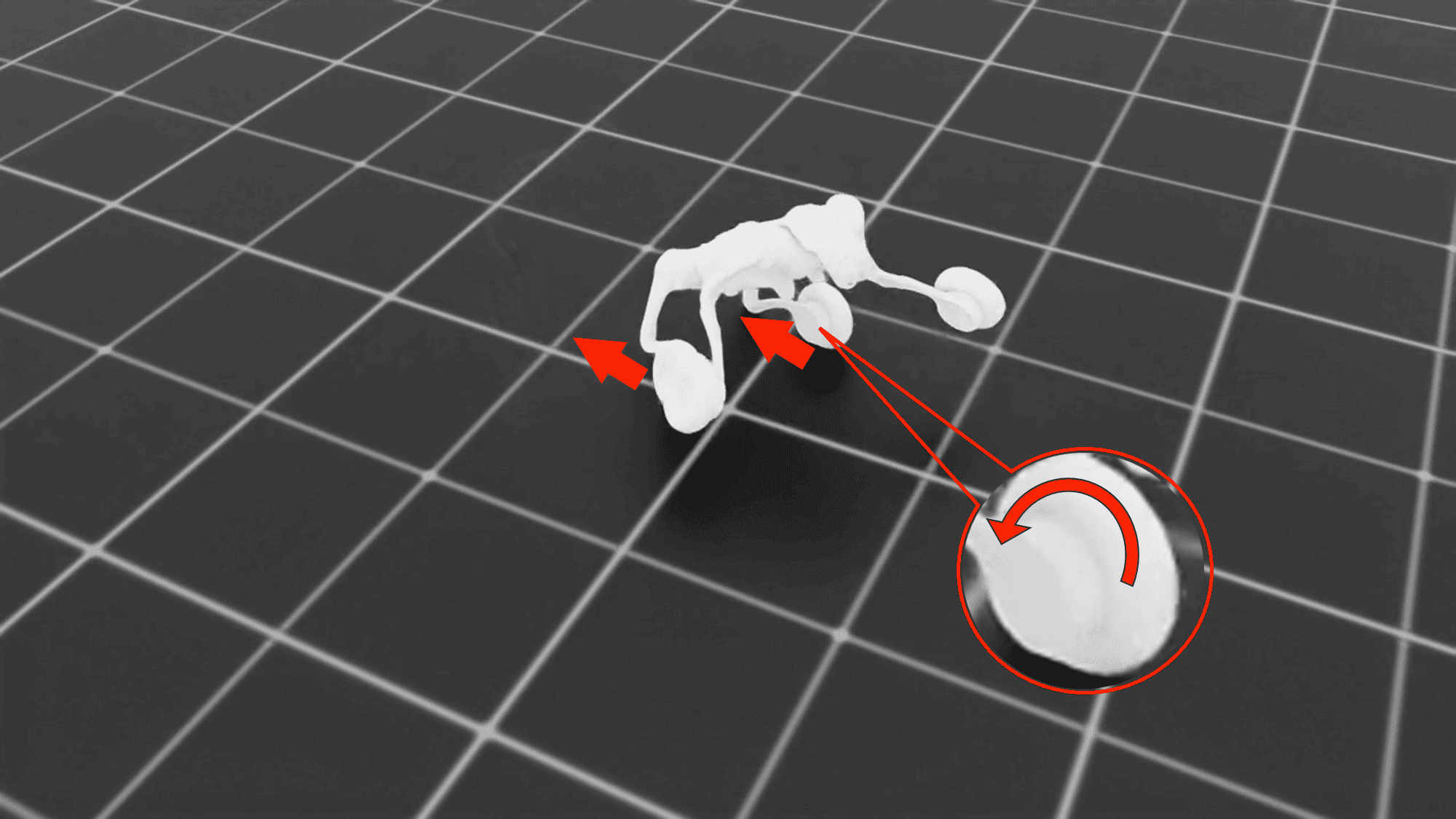}
        \caption{}
    \end{subfigure}%
    \vspace{0.1pt}
    \\
    \begin{subfigure}[b]{0.15\textwidth}
        \centering
        \includegraphics[trim=5cm 0 3cm 2,clip,width=\textwidth,height=0.8\textwidth,keepaspectratio=false]{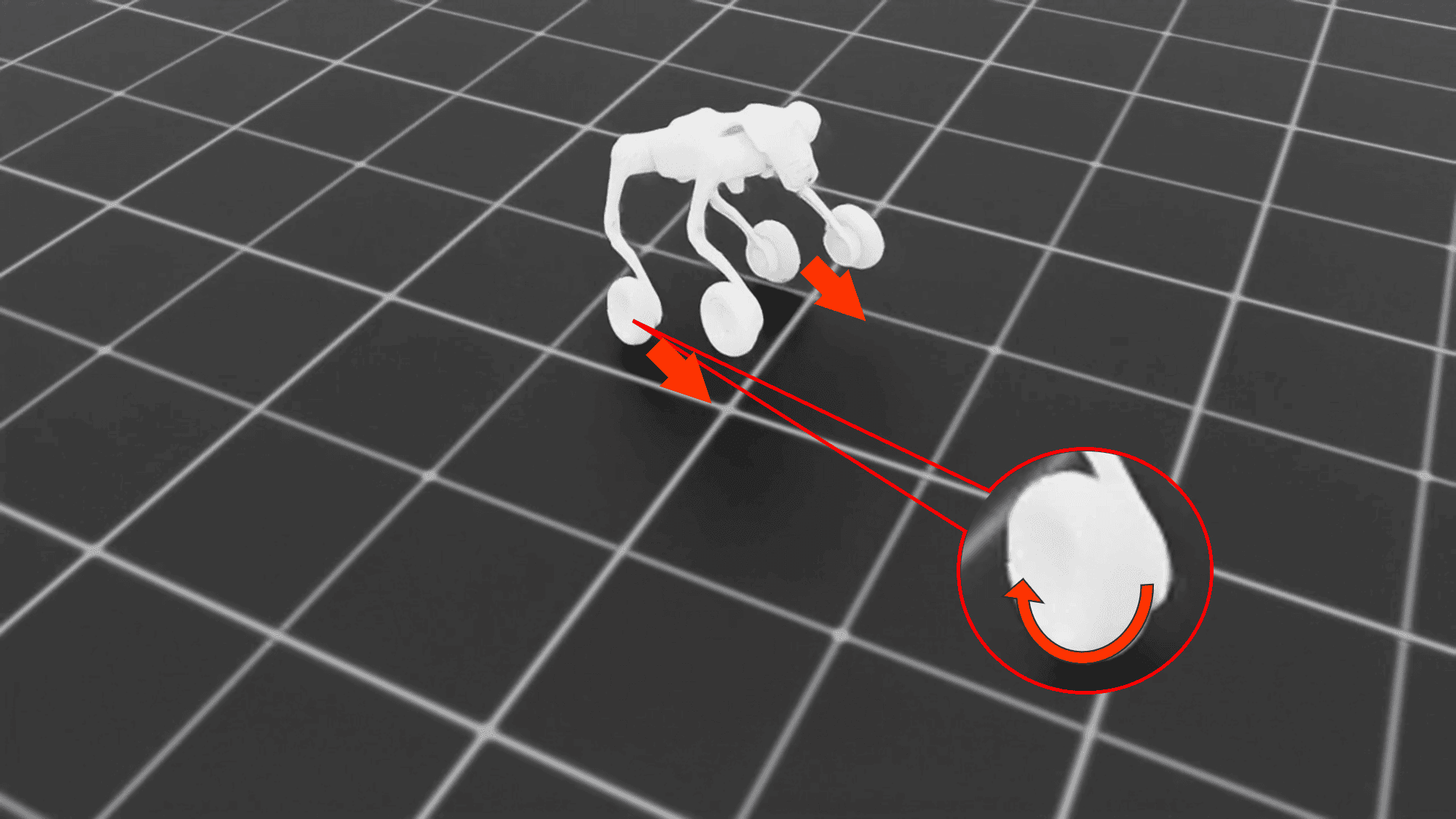}
        \caption{}
    \end{subfigure}%
    \hspace{0.1cm}
    \begin{subfigure}[b]{0.15\textwidth}
        \centering
\includegraphics[trim=4cm 0 4cm 2,clip,width=\textwidth,height=0.8\textwidth,keepaspectratio=false]{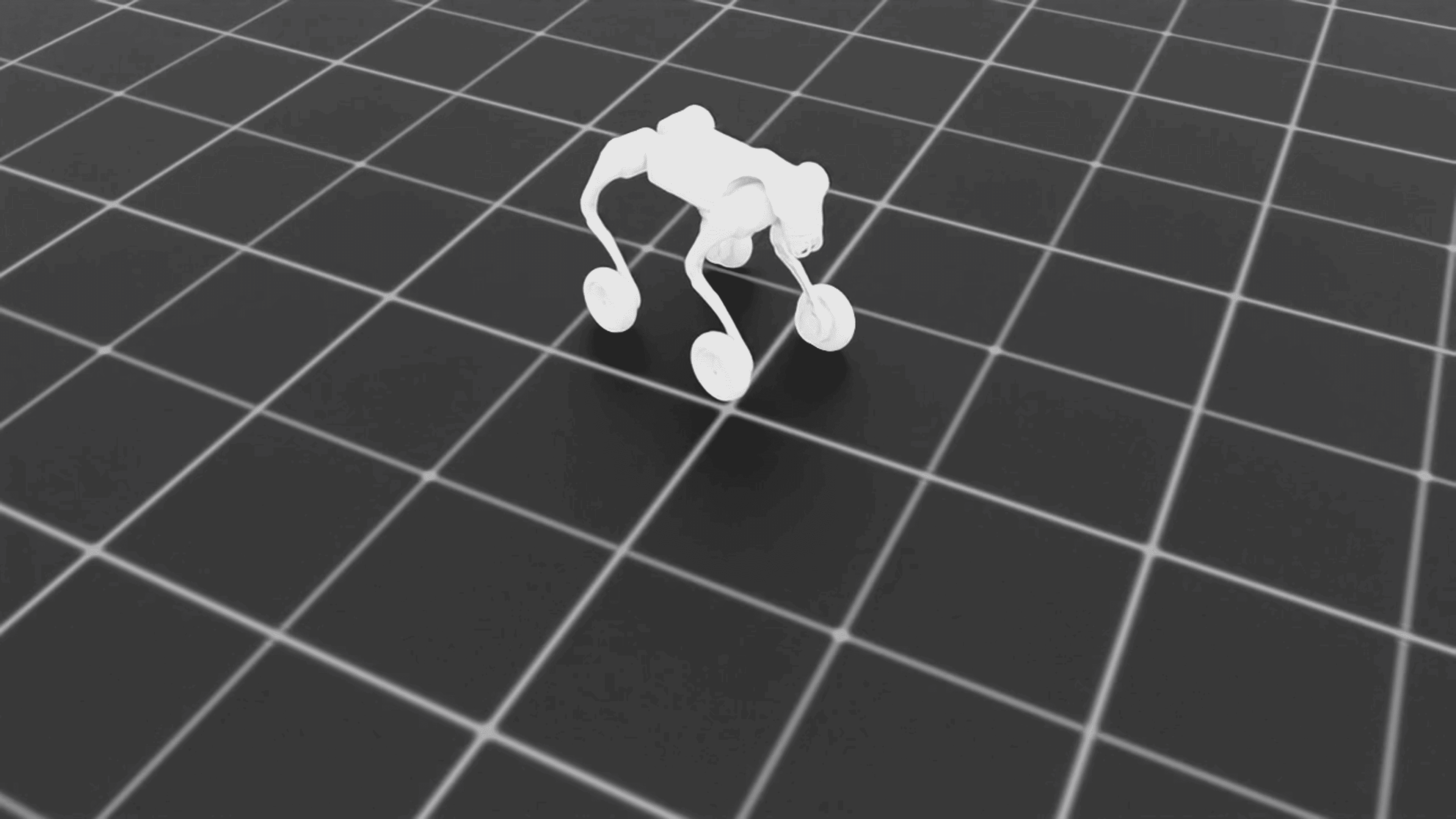}
        \caption{}
    \end{subfigure}
    \caption{Key stages of Unitree Go2-W\cite{Unitree} ’s recovery process from prone to standing.In (b), after standing up, the robot tilts backward, and backward wheel rotation buffers the tilt to prevent tipping over. Once the center of mass is stabilized in (c), forward wheel rotation quickly adjusts the base height and joint configuration.}
    \label{unitree}
\end{figure}

\begin{table}[H]
\centering
\caption{Comparison of Recovery Success Rates}
\begin{tabular}{cccc}
\hline
Platform       & Strategy & Success Rate (\%) \\ \hline
\multirow{2}{*}{KYON}  & ED       & \textbf{99.1}              \\
               & Baseline & 96.4              \\ \hline
\multirow{2}{*}{Unitree Go2-W\cite{Unitree}} & ED & \textbf{97.8}              \\
                      & Baseline & 94.1       \\ \hline
\end{tabular}
\label{table suc rate}
\end{table}
\begin{table}[H]
\centering
\caption{Comparison of Average Joint Torque}
\resizebox{\columnwidth}{!}{
\begin{tabular}{cccc}
\hline
Platform       & Configuration & Avg Torque ($N \cdot m$) & Reduction (\%) \\ \hline
KYON  & Wheel Active      & 35.776            &      \\
Max Torque:132  & Wheel Fixed  & 42.515     & \textbf{15.85}      \\ \hline
Unitree Go2-W\cite{Unitree} & Wheel Active & 7.123     &          \\
Max Torque:23.7  & Wheel Fixed    & 9.657    & \textbf{26.24}        \\ \hline
\end{tabular}
}
\label{table torque com}
\end{table}
\subsection{Recovery under Repeated Perturbations}
\begin{figure*}[!t]
    \centering
        \centering
        \begin{subfigure}[b]{0.18\textwidth}
            \centering
            \includegraphics[width=\textwidth]{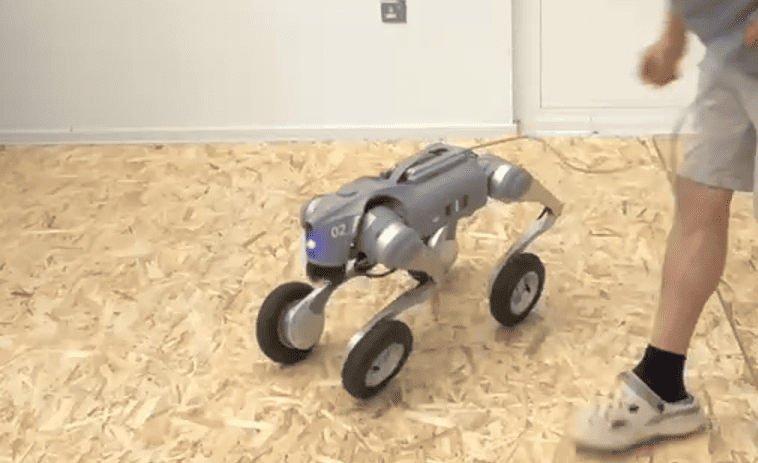}
            \caption{}
        \end{subfigure}
        \hspace{1pt}
        \begin{subfigure}[b]{0.18\textwidth}
            \centering
            \includegraphics[width=\textwidth]{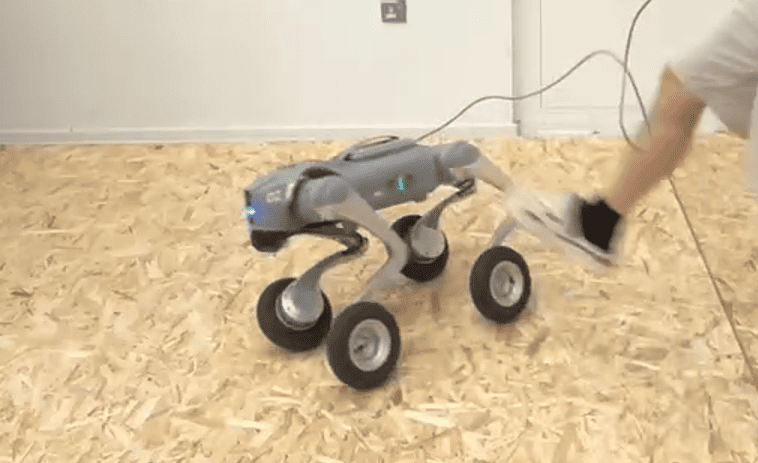}
            \caption{}
        \end{subfigure}
        \hspace{1pt}
        \begin{subfigure}[b]{0.18\textwidth}
            \centering
            \includegraphics[width=\textwidth]{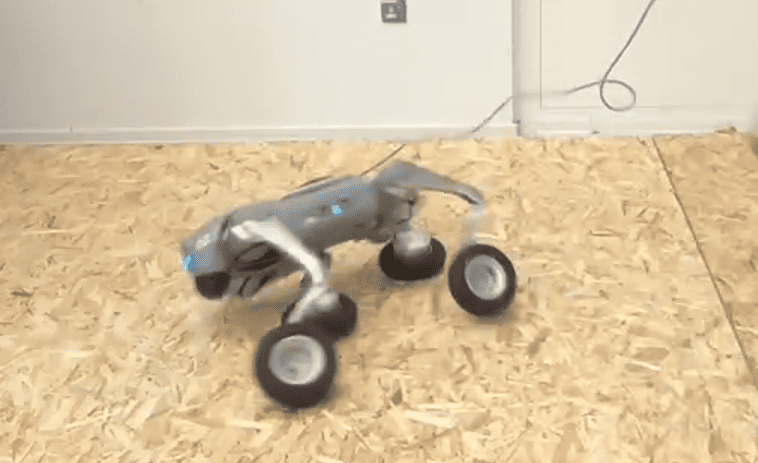}
            \caption{}
        \end{subfigure}
        \hspace{1pt}
        \begin{subfigure}[b]{0.18\textwidth}
            \centering
            \includegraphics[width=\textwidth]{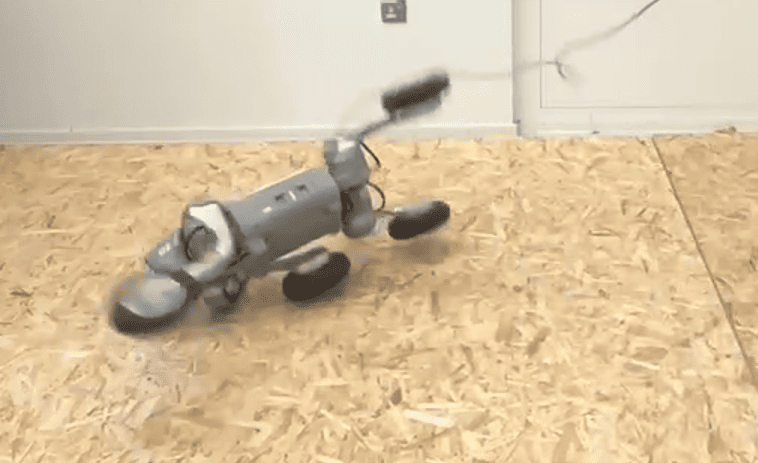}
            \caption{}
        \end{subfigure}
        \hspace{1pt}
        \begin{subfigure}[b]{0.18\textwidth}
            \centering
            \includegraphics[width=\textwidth]{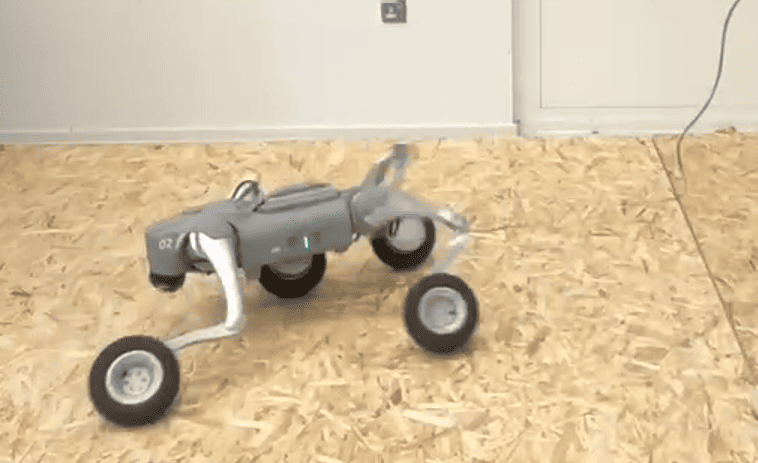}
            \caption{}
        \end{subfigure}
        \vskip 0.5em
        \begin{subfigure}[b]{0.18\textwidth}
            \centering
            \includegraphics[width=\textwidth]{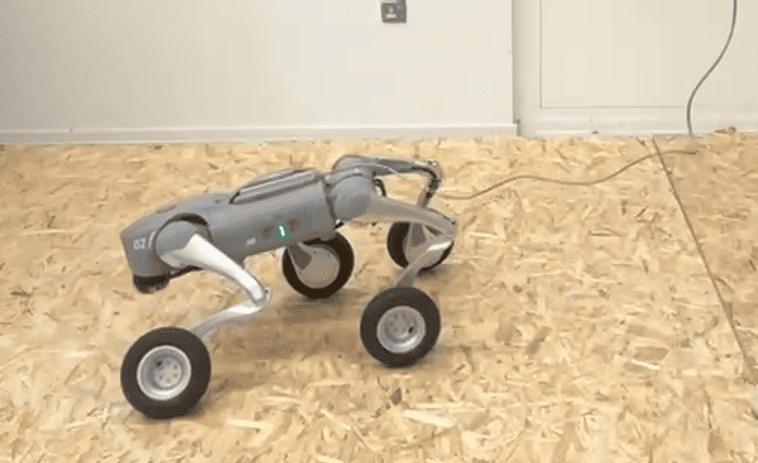}
            \caption{}
        \end{subfigure}
        \hspace{1pt}
        \begin{subfigure}[b]{0.18\textwidth}
            \centering
            \includegraphics[width=\textwidth]{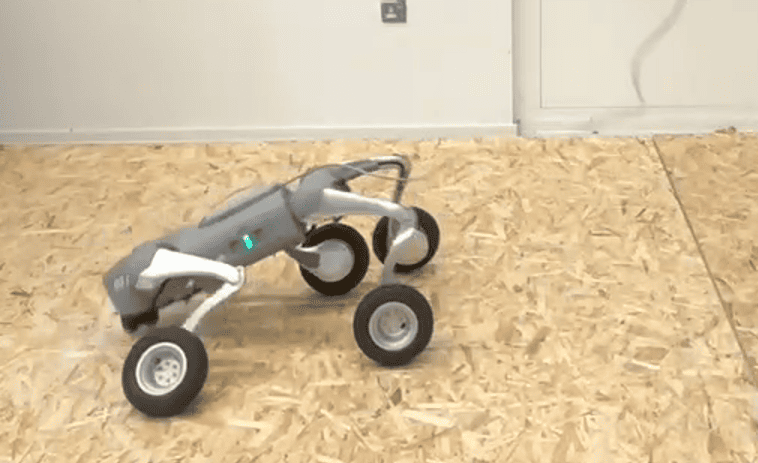}
            \caption{}
        \end{subfigure}
        \hspace{1pt}
        \begin{subfigure}[b]{0.18\textwidth}
            \centering
            \includegraphics[width=\textwidth]{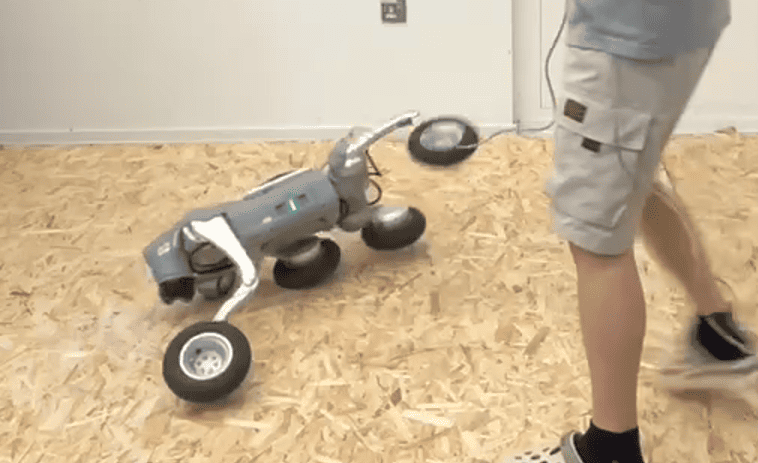}
            \caption{}
        \end{subfigure}
        \hspace{1pt}
        \begin{subfigure}[b]{0.18\textwidth}
            \centering
            \includegraphics[width=\textwidth]{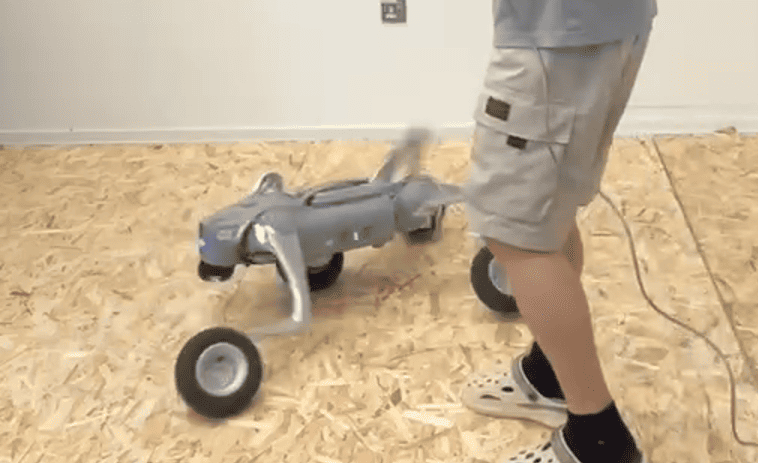}
            \caption{}
        \end{subfigure}
        \hspace{1pt}
        \begin{subfigure}[b]{0.18\textwidth}
            \centering
            \includegraphics[width=\textwidth]{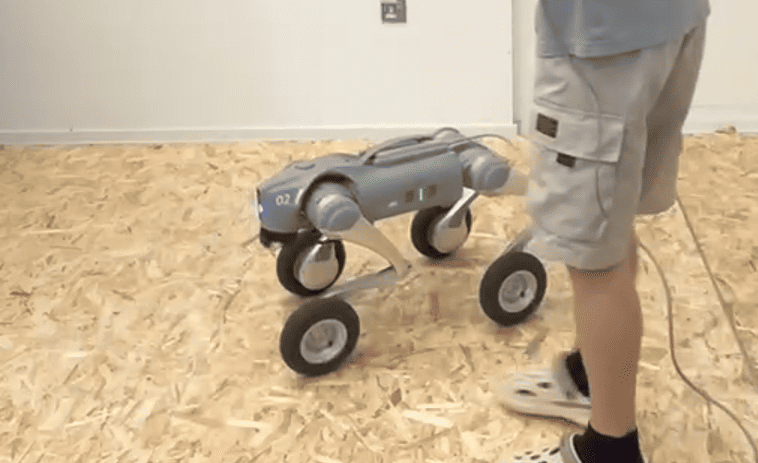}
            \caption{}
        \end{subfigure}
    \caption{Recovery under Repeated Perturbations: (b)-(c) externally induced second fall; (d)–(g) unstable stand-up attempt leading to a re-fall; (h)–(j) autonomous second recovery.The Video is provided on the website.}
    \label{fig:Recovery under Repeated Perturbations}
\end{figure*}

To accelerate training, we adopted a fixed-step recovery setting in simulation. In real-world tests, once the robot reached a standing posture, we applied multiple manual perturbations without reset. The policy autonomously recovered after each perturbation, Fig. \ref{fig:Recovery under Repeated Perturbations} (a)–(g); when the initial recovery was insufficient, it immediately corrected the motion and performed a second recovery, Fig. \ref{fig:Recovery under Repeated Perturbations} (h)–(j). These results indicate that, at test time, recovery is sustained and not constrained by the fixed step used during training.
\subsection{Recovery Adaptability on Unseen Terrains}
To investigate the emergent terrain adaptation capabilities of our method, we also evaluated the policy on five procedurally generated, uneven terrains:\begin{enumerate}
    \item Random Boxes: Discontinuous platforms with $0.05m$-$0.2 m$ height variations.
    \item Rough Terrain: High-frequency unevenness ($\delta=0.02m$-$0.10 m$).
    \item Sloped Pyramid: Inclined surfaces with $20\%$–$60\%$ gradients.
    \item Pyramid Stairs: Ascending/descending stairs with step heights $0.05m$–$0.23m$.
    \item Inverted Pyramid Stairs: The opposite of Pyramid Stairs.
\end{enumerate}
Notably, the policy exhibits partial zero-shot generalization: the robot maintains stability through controlled wheel-leg coordination, despite the absence of explicit terrain sensing. We compared the recovery success rates of the ED strategy and the baseline strategy deployed on the KYON robot in non-flat environments, defining recovery as having a joint angle deviation from the default configuration of no more than $0.5rad$ and a base orientation error—calculated as the sum of differences between the robot’s projected gravity vector and the ideal direction $\left [ 0,0,-1 \right ] $ —of less than $0.1$, with the success rates shown in the Table \ref{table suc rate on non-flat}.

However, recovery success rates decline significantly under these conditions, with frequent secondary falls observed—particularly on stair terrains. These undesirable behaviors, which undermine real-world deployment, can be attributed to two unresolved challenges. First, the policy’s reactive adjustments lack explicit perception of terrain geometry, causing delayed or inadequate responses when contact conditions change abruptly such edge or step transitions. Second, the learned posture adaptation remains largely configuration-driven rather than terrain-aware, resulting in suboptimal body alignment and insufficient stability margins on uneven or sloped surfaces.
\begin{table}[H]
\centering
\caption{Comparison of Recovery Success Rates on non-flat terrain}
\begin{tabular}{cccc}
\hline
Platform       & Strategy & Success Rate (\%) \\ \hline
\multirow{2}{*}{KYON}  & ED       & \textbf{78.6}              \\
               & Baseline & 61.8              \\ \hline
\end{tabular}
\label{table suc rate on non-flat}
\end{table}
\begin{figure}[t]
    \centering
    \begin{subfigure}[b]{0.18\textwidth}
        \centering
        \includegraphics[width=\textwidth]{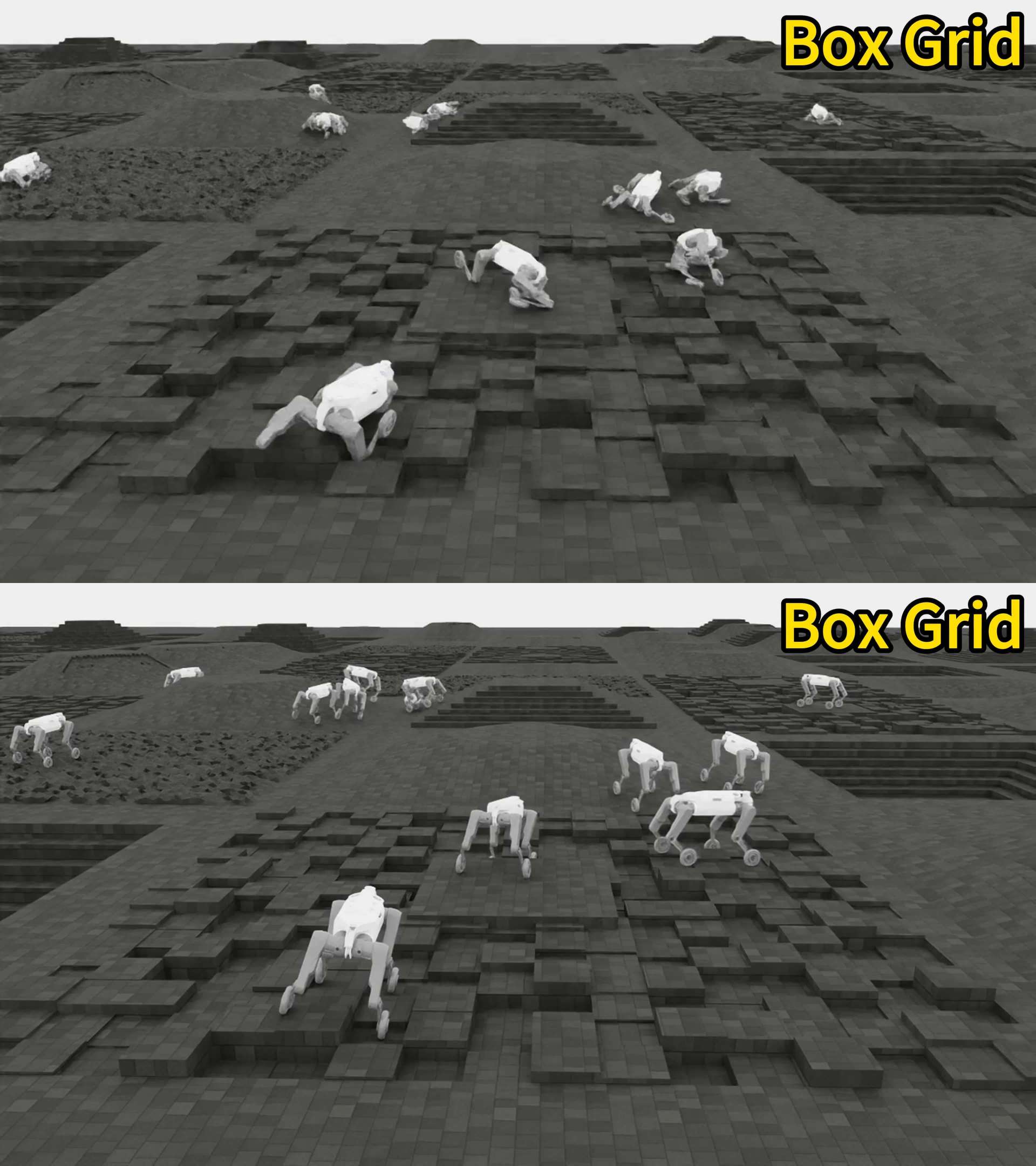}
        \caption{}
    \end{subfigure}
    \hspace{1pt}
    \begin{subfigure}[b]{0.18\textwidth}
        \centering
        \includegraphics[width=\textwidth]{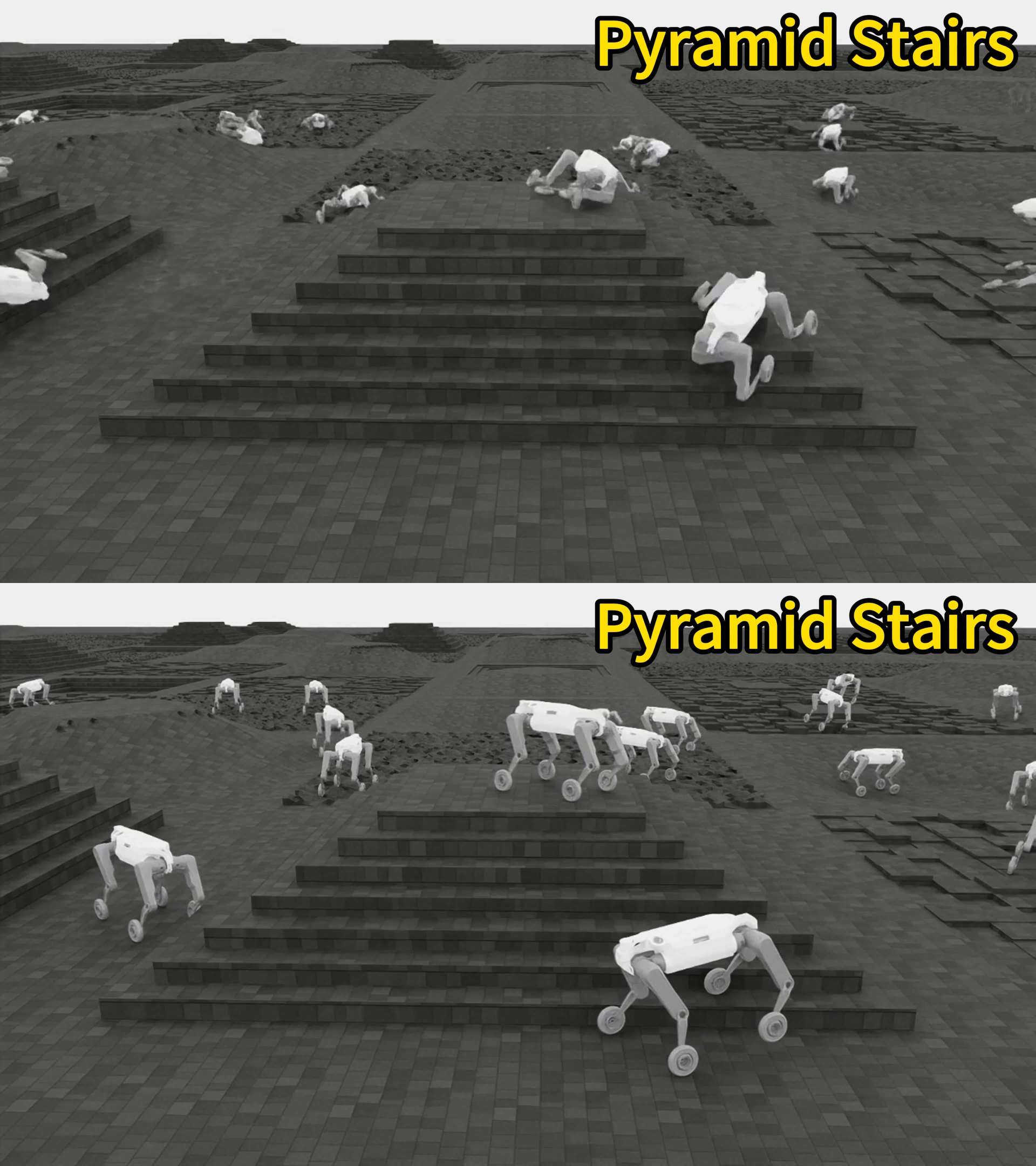}
        \caption{}
    \end{subfigure}
    \caption{Evaluation of the ED-policy on the KYON Robot in irregular terrain.}
    \label{fig:nonflat}
\end{figure}
\section{CONCLUSION}
This paper presents a Episode-based Dynamic Reward Shaping for wheel-legged robots to achieve robust and adaptive recovery from fallen states. By combining dynamic reward shaping and curriculum learning, the proposed method effectively overcomes the exploration-convergence dilemma inherent in sparse-reward reinforcement learning. Experimental results demonstrate that ED achieves a recovery success rate of over $97\%$ across diverse platforms, outperforming baseline methods by 3 percentage points. The core innovation stems in the synergy between wheel-assisted motion and leg articulation: wheels facilitate rapid centroidal adjustments via controlled rolling, while legs ensure fine-grained posture stabilization., collectively reducing joint torque demands by $15-26\%$ compared to fixed-wheel configurations. Cross-platform validation on robots with distinct kinematic and dynamic parameters (KYON and Unitree Go2-W\cite{Unitree}) further confirms the generalization capability of the framework, where ED autonomously adapts to hardware-specific constraints without manual tuning. These findings highlight the potential of wheel-leg coordination as a universal strategy for recovery control in hybrid locomotion systems.

Future work will focus on extending the framework to handle more complex terrains (e.g., slopes, gravel) and integrating real-world sensor noise models to bridge the sim-to-real gap. Moreover, the principles of dynamic reward shaping could be generalized to other discontinuous-contact tasks, including multi-robot collaborative recovery, manipulation-oriented posture adjustment, and adaptive locomotion in cluttered environments, potentially enabling more robust, versatile, and energy-efficient robotic behaviors.
\section*{ACKNOWLEDGMENT}

We would like to thank Andrea Patrizi, 
Despoina Maligianni, Rui Dai, Yifang Zhang, Maolin Lei, Jingcheng Jiang, Kuanqi Cai, and Carlo Rizzardo for their useful discussions.

\section*{References Section}

\renewcommand{\refname}{}
\bibliographystyle{IEEEtran}
\bibliography{ref}
\end{document}